\pdfoutput=1
\documentclass[11pt]{article}
\usepackage[margin=1.1in]{geometry}
\usepackage{amsmath,amssymb,amsthm,mathtools}
\usepackage{graphicx}
\usepackage{booktabs}
\usepackage[dvipsnames,table]{xcolor}
\usepackage{microtype}
\usepackage[numbers,sort&compress]{natbib}
\usepackage[colorlinks=true,linkcolor=blue!60!black,citecolor=blue!60!black,urlcolor=blue!60!black]{hyperref}
\usepackage{subcaption}

\newtheorem{theorem}{Theorem}[section]
\newtheorem{proposition}[theorem]{Proposition}
\newtheorem{lemma}[theorem]{Lemma}
\newtheorem{corollary}[theorem]{Corollary}

\newcommand{\errBest}{4.55}          % best high-data pipeline test error (%)
\newcommand{\errBestLow}{5.38}       % best low-data pipeline test error, N=1250 (%)
\newcommand{\errMLP}{4.86}           % MLP + TTA, high data
\newcommand{\errRef}{4.73}           % kernel-conditioned refiner + TTA
\newcommand{\errMSE}{4.71}           % MSE-trained MLP + TTA
\newcommand{\errFNOG}{4.70}          % FNO member + TTA
\newcommand{\errUNetG}{4.99}         % UNet member + TTA
\newcommand{\errStackFam}{4.67}      % MLP-family stack (before diverse members)
\newcommand{\errStack}{4.58}          % diverse per-pixel stack before correction
\newcommand{\uqDisAbsDiv}{0.08}      % corr(disagreement, abs err), diverse ensemble
\newcommand{\corrNetLo}{0.86}        % min network-network residual corr
\newcommand{\corrNetHi}{0.96}        % max network-network residual corr
\newcommand{\corrKerLo}{0.79}        % min network-kernel residual corr
\newcommand{\corrKerHi}{0.89}        % max network-kernel residual corr
\newcommand{\errKRRload}{5.19}       % kernel ridge on loads, high data
\newcommand{\errKRRloadLow}{6.66}    % multiscale kernel ridge on loads, N=1250
\newcommand{\normRatio}{271}         % RKHS norm ratio direct/residual
\newcommand{\uqCoverNom}{90}         % nominal conformal level (%)
\newcommand{\uqCover}{91.6}          % empirical conformal coverage (%)
\newcommand{\uqCorrAbs}{0.08}        % corr(power fn, absolute error) -- weak
\newcommand{\uqCorrNorm}{0.65}       % corr(power fn, output norm) -- the confound
\newcommand{\uqDisAbs}{0.10}         % corr(ensemble disagreement, absolute error) -- also weak
\newcommand{\pubBestHi}{4.55}        % best published high-data (PARA-Net)
\newcommand{\pubBestLo}{6.49}        % best published low-data (FNO-mean GP)
\newcommand{\ttaGainMLP}{0.11}       % MLP test-time reflection-averaging gain (points)

\title{Neural means and kernel corrections for operator learning}
\author{%
Yitzchak Shmalo%
\thanks{Einstein Institute of Mathematics, The Hebrew University of Jerusalem,
Jerusalem, Israel. \texttt{yitzchak.shmalo@gmail.com}.
Code, data pointers, and the per-run summaries behind every number reported here are at
\url{https://github.com/yspennstate/neural-means-kernel-corrections}.}}
\date{July 2026}

\begin{document}
\maketitle

\begin{abstract}
We combine neural network means with exact Mat\'ern kernel regressions of
their residuals and of their learned features, and evaluate the pairing on two
public emulation problems with published baselines: the structural-mechanics
benchmark of de Hoop et al.\ and the OCO-2 radiative-transfer emulator of
Lamminp\"a\"a et al. On structural mechanics the combination reaches
\errBest\% test error, matching the best published architecture, and
\errBestLow\% against a published \pubBestLo\% in the low-data regime. On
OCO-2 it improves on the published Gaussian-process emulator on that problem's
own test points, outright on two of the three spectral bands; the same kernel
that trails the network tenfold on the raw state overtakes it on the network's
features, and we measure why (the target's squared native-space norm drops
about fortyfold at fixed effective dimension) and prove the mechanism. Where
the two families tie instead, the residuals of every architecture we train
correlate above 0.86 and their shared component is flat in diversity and
sample size, which reads the published plateau as a property of the data.
Supporting results include a second-moment identity that predicts stacking
outcomes from measured correlations, an optimal-recovery certificate, and a
distribution-free coverage band, the only uncertainty signal that survives our
tests.
\end{abstract}

\section{Introduction}
\label{sec:intro}
% written after results freeze -- see draft in intro.tex include
Operator learning constructs fast surrogates for the solution maps of
parametric partial differential equations and physical forward models.
Neural operators \citep{kovachki2023neural,lu2021deeponet,li2021fourier}
learn their own representations; kernel and Gaussian-process methods
\citep{batlle2024kernel,mora2025operator,owhadi2019operator} come with exact
solves and error certificates and a strong record on the same benchmarks.
This paper combines the two: a neural network mean, trained in the metric the
application reports, with an exact Mat\'ern kernel regression of its residual
and of its learned features. We evaluate the combination in depth on two
public problems with published baselines --- the structural-mechanics
benchmark of \citet{dehoop2022cost} and the OCO-2 radiative-transfer
emulation problem of \citet{susiluoto2025radiative} --- and, with one
unchanged recipe, across the seven-benchmark operator-learning suite
assembled by \citet{batlle2024kernel} from the problems of
\citet{dehoop2022cost} and \citet{lu2022comprehensive}.
Table~\ref{tab:suite} collects the published numbers and ours.

\begin{table}[t]
\centering
\caption{The two problems studied in depth (OCO-2, structural mechanics) and
the rest of the operator-learning suite, with the best published mean
relative $L^2$ test error and ours. OCO-2 rows compare against the
Gaussian-process emulator of \citet{susiluoto2025radiative} on its own test
points, in that paper's reduced-coefficient metric; Section~\ref{sec:oco2}
reports both of its metrics, on which our O2 and strong-CO$_2$ surrogates win
both while the weak-CO$_2$ metrics are won by two different heads. The
published suite numbers are per-problem methods tuned by their authors; ours
outside the two problems studied in depth come from one recipe run unchanged
(a Mat\'ern kernel at the median length scale, fit capped at 6000 points, a
residual MLP mean, outputs reduced by PCA, the stage selected on validation).
Burgers and Darcy use slightly smaller training splits than their baselines
(800 and 824 pairs) and Darcy a different grid, from the copies we could
obtain.}
\label{tab:suite}
\begin{tabular}{llcc}
\toprule
Problem & Map & Best published & This work \\
\midrule
OCO-2, O2 band & atmospheric state $\to$ radiance & 16.89\% & 3.83\% \\
OCO-2, weak CO$_2$ & atmospheric state $\to$ radiance & 24.1\% & 16.1\% \\
OCO-2, strong CO$_2$ & atmospheric state $\to$ radiance & 16.14\% & 7.96\% \\
Structural mechanics & boundary load $\to$ stress & 4.55\% (PARA-Net) & 4.55\% \\
\midrule
Burgers & initial condition $\to$ solution & 1.93\% (FNO) & 2.38\% \\
Darcy flow & permeability $\to$ pressure & 2.32\% (POD-DeepONet) & 2.01\% \\
Advection~I & smooth pulse $\to$ solution & $\approx0\%$ (kernel) & --- \\
Advection~II & discontinuous pulse $\to$ solution & 11.28\% (linear kernel) & 11.29\% \\
Helmholtz & wavespeed $\to$ field & 1.00\% (kernel) & 1.75\% \\
Navier--Stokes & vorticity $\to$ vorticity & 0.12\% (kernel) & 0.51\% \\
\bottomrule
\end{tabular}
\end{table}

The two problems studied in depth sit at opposite ends of the
neural-versus-kernel spectrum, and the pair is the point. On structural
mechanics the families tie: the pipeline reaches \errBest\% under the
20000-sample protocol, matching the best published architecture (PARA-Net,
\pubBestHi\%) within run-to-run noise, and \errBestLow\% under the
1250-sample protocol against a published best of \pubBestLo\%. What the
coupling adds there is mostly understanding. The residuals of every
architecture we train correlate between \corrNetLo{} and \corrNetHi{}, their
shared component concentrates where the finite element data are least
reliable and is flat in ensemble diversity and training-set size, and so the
evidence reads the published plateau near 4.5\% as a property of the
benchmark's data rather than of any surrogate class
(Section~\ref{sec:diversity}). Absent regenerated finite-element data this
remains an inference, but it is the reading every measurement we made
supports. On the rest of the suite the same recipe sorts the problems by the
same logic: where the map is smooth and data are plentiful (Helmholtz,
Navier--Stokes, Darcy) the kernel stage wins the pipeline's internal
selection and lands within a factor of two of the tuned per-problem kernels
of \citet{batlle2024kernel}, and on Burgers the corrected mean sits near the
Fourier neural operator.

On OCO-2 the same components separate by an order of magnitude, and the
pipeline's job changes: not to couple two comparable models but to move the
kernel's exactness onto the network's representation. Per spectral band of
the satellite, the task maps a reduced atmospheric state to a reduced
radiance spectrum, and the baseline is the kernel-flow emulator of
\citet{susiluoto2025radiative}, whose stored predictions we score on the same
test points. An exact Mat\'ern head on the trained network's features reaches
$3.8\%$ where the same kernel on the raw state reaches $40\%$; the effective
dimension of the two Gram matrices is unchanged, the target's squared
native-space norm falls by a factor of about forty, and
Proposition~\ref{prop:pullback} gives the mechanism. Trained in the reported
metric and combined per output coordinate, the result improves on the
emulator on both of that problem's error metrics at once on two of the three
bands; on the third each metric is won by a separate head
(Section~\ref{sec:oco2} is precise about which).

Each stage of the pipeline carries one supporting result: a symmetrization
lemma behind the reflection averaging, a second-moment identity that predicts
stacking outcomes from measured residual correlations, an optimal-recovery
bound $\|G-m\|_K\,P_\lambda(u)$ for the corrected surrogate, a pullback
identity and a rate for the feature kernel, and a distribution-free coverage
statement for the reported uncertainty. The last one matters because the
design factor $P_\lambda$, though a valid bound, turns out to be a poor
pointwise ranking of the actual error, and its split-conformal rescaling is
the only uncertainty signal that survives our tests (Section~\ref{sec:uq}).

Section~\ref{sec:benchmark} describes the problems, protocols, and published
results. Section~\ref{sec:method} specifies the pipeline.
Sections~\ref{sec:experiments} and~\ref{sec:oco2} report the two studies,
including the data-scaling experiments. Section~\ref{sec:theory} states the
supporting theory, one result per stage, with proofs in
Appendix~\ref{app:proofs}, and Section~\ref{sec:discussion} discusses
limitations.

\section{Benchmark, protocol, and related work}
\label{sec:benchmark}
\subsection{Problem and data}
The dataset originates in the cost-accuracy study of \citet{dehoop2022cost}
and is the one distributed with \citet{batlle2024kernel} and
\citet{mora2025operator}. An isotropic elastic plate occupies
$D=(0,1)^2$; the displacement field $w$ satisfies
$\nabla\cdot\sigma=0$ with a fixed constitutive law, displacement conditions
on the bottom and lateral parts of the boundary, and a prescribed normal
traction $\bar t$ on the top edge $\Gamma_t=[0,1]\times\{1\}$. The quantity of
interest is the von Mises stress field $\sigma_v$ on $D$. The learning task is
the map $G:\bar t\mapsto\sigma_v$. Loads are drawn from the Gaussian field
$\mathcal{GP}\big(100,\,400^2(-\Delta+3^2 I)^{-1}\big)$ with homogeneous
Neumann boundary conditions for the Laplacian; outputs are finite element
solutions interpolated to a regular $41\times41$ grid, and the load is sampled
at $41$ points. The distributed file contains 40000 input/output pairs; the
input array stores the load broadcast along the second grid coordinate, which
we verified is an exact copy (maximal deviation $0$ across all 40000
samples), so all our methods consume the load as a vector in $\mathbb R^{41}$.

Errors are mean relative $L^2$ errors over the test set,
\[
\frac1{N_{\mathrm{test}}}\sum_{n}\frac{\|\widehat
G(u_n)-G(u_n)\|_2}{\|G(u_n)\|_2},
\]
computed on grid values in double precision. Quadrature weighting
(trapezoidal instead of plain vector norms) changes our reported numbers by
less than 0.02 percentage points, and we report the plain-norm convention of
the prior work.

\subsection{Protocols and published results}
Two regimes appear in the literature, and we follow both. In the
\emph{high-data} protocol the first 20000 samples are available for training
and the last 20000 form the test set; \citet{dehoop2022cost} report, at 20000
training samples, 4.55\% (PARA-Net), 4.67\% (PCA-Net), 4.76\% (FNO) and 5.20\%
(DeepONet), and \citet{batlle2024kernel} report 5.18\% for their
optimal-recovery kernel method (Mat\'ern/rational quadratic; 27.11\% for the
linear kernel) on the same task. Within the 20000-sample budget we hold out
the last 1000 samples (of a fixed permutation) for model selection and
stacking, and train on the remaining 19000, so no method of ours sees more
than the 20000 training samples used by the baselines. In the \emph{low-data}
protocol of \citet{mora2025operator}, 1250 samples are available and the same
20000-sample test set is used; their table gives 8.70\% (DeepONet), 6.62\%
(FNO), 6.95\% (the kernel method of \citealt{batlle2024kernel}), 6.74\% (their
zero-mean GP), 7.12\% (DeepONet-mean GP) and 6.49\% (FNO-mean GP). In this
regime we carve the validation split (250 samples) out of the 1250, so
training uses 1000 samples and every choice made by the pipeline is informed
by the 1250 available labels only.

\subsection{A data-driven check of the mirror symmetry}
\label{sec:symcheck}
The continuous problem is invariant under $x_1\mapsto1-x_1$: the domain and
boundary partition are symmetric, and the input law has a symmetric
covariance. On the grid, reflecting the load ($S$) should reflect the stress
field along the first grid coordinate ($T$), i.e.\ $G(Su)=TG(u)$. Rather than
assume this, we test it on the data. For each of the first 200 samples we
searched all 40000 loads for the nearest neighbor of the reflected load $Su_i$
and compared the corresponding outputs. For the five best-matching pairs, the
input mismatch $\|Su_i-u_j\|/\|u_j\|$ ranges over $0.27$--$0.32$, while the
output mismatch $\|Tv_i-v_j\|/\|v_j\|$ ranges over $0.085$--$0.14$; reflecting
along the second coordinate instead gives output mismatches of order one.
Outputs of near-mirror inputs are far closer than the inputs themselves,
which is what equivariance plus a Lipschitz solution map predicts, and the
effect singles out the correct output reflection axis. A complementary check
on the input law: the empirical mean and covariance of the training loads are
invariant under reflection to within 1.1\% and 1.5\% respectively, as required
for Proposition~\ref{prop:sym}. Consistently with all of this, reflection
averaging at test time improves every model we trained
(Section~\ref{sec:experiments}), as Proposition~\ref{prop:sym} says it must on
average when the symmetry holds.

\subsection{Related work}
\label{sec:related}
The benchmark sits at the meeting point of three lines of work. \emph{Neural
operators} learn maps between function spaces: DeepONet \citep{lu2021deeponet}
with its branch/trunk factorization, the Fourier neural operator
\citep{li2021fourier} and the broader neural-operator framework
\citep{kovachki2023neural}, and the reduced-basis PCA-Net and PARA-Net
architectures assembled for the cost--accuracy study of
\citet{dehoop2022cost}. These methods are fast and mesh-flexible but do not by
themselves come with error control, and it is their numbers on this problem
that define the plateau we start from. \emph{Kernel and Gaussian process
methods} for operator learning approach the same maps through reproducing
kernels: the optimal-recovery framework of \citet{batlle2024kernel}, with its
convergence theory and a-priori bounds, is the most direct comparison, and it
is competitive with neural operators on most of the benchmarks it considers.
Data-adapted kernels learned by cross-validation \citep{darcy2023one} and
vector-valued kernel formulations extend the reach of this family. \emph{Hybrids}
that place a neural network and a kernel in the same estimator are the closest
relatives of our pipeline: \citet{mora2025operator} use a neural operator as
the mean of a Gaussian process and fit the two together, and this is the work
we most directly build on, differing in that we fit the mean first and the
kernel after, tune by cross-validation rather than marginal likelihood, and
solve the correction exactly at nineteen thousand points.

Two further threads inform the design. The kernel-flow method
\citep{owhadi2019kernel} learns a kernel from data by minimizing a
half-sample cross-validation loss, and its use as a regularizer for the inner
layers of a network \citep{yoo2021deep} is exactly the term we analyze in
Proposition~\ref{prop:kf}; kernel flows have since been used at scale as
emulators for physical forward models, for instance in atmospheric radiative
transfer retrievals \citep{susiluoto2025radiative} and in the inference of
convective-storm structure from satellite observations
\citep{prasanth2021kernel}, where presenting the input in its physical
parametrization and adapting the kernel to data are what make the emulator
accurate. Our pipeline follows the same instinct, with the smoothing of the
target performed by a neural ensemble rather than by a hand-chosen reduction.
Finally, the effective-dimension reading of the kernel solve
(Lemma~\ref{lem:deff}) draws on the random-matrix description of kernel Gram
spectra \citep{koltchinskii2000random} and on the classical
Mar\v{c}enko--Pastur \citep{marchenko1967distribution} and spiked-covariance
\citep{baik2005phase} laws; the same effective dimension governs the
generalization of kernel ridge regression \citep{caponnetto2007optimal}.

\section{Method}
\label{sec:method}
The pipeline has three stages: neural means, stacking, and kernel
corrections. All components operate on the load vector $u\in\mathbb R^{41}$
(standardized by training statistics) and produce stress fields in
$\mathbb R^{41\times41}$; all networks are trained with the reported metric as
the loss, i.e.\ the per-sample relative $L^2$ error in original units, which
we found mildly but consistently better than mean squared error on
normalized targets.

\subsection{Neural means}
\emph{Residual MLP.} The primary mean is a residual multilayer perceptron:
three or five residual SiLU blocks of width 1024--1536 mapping
$\mathbb R^{41}\to\mathbb R^{1681}$ directly. This is essentially PARA-Net
\citep{dehoop2022cost} with a modern training recipe, and it also supplies the
penultimate features used by the feature-space kernel stage.

\emph{Kernel-conditioned refiner.} The second mean is the same residual network
given an extra input channel: the kernel method's predicted stress field for
the same load, concatenated with the load. It therefore learns to correct the
kernel predictor rather than to reproduce the map from scratch, and it is the
strongest single model in our study. During training the refiner reads
\emph{out-of-fold} kernel predictions (four-fold, so the kernel channel is
never fit on the target sample), and at test time the full-data kernel
prediction; this keeps the channel honest.

\emph{Other instances.} The mean slot is not tied to a particular
architecture. We also use a Fourier neural operator \citep{li2021fourier}
that consumes the broadcast load as a field, a UNet on the same field
representation, and an MSE-trained variant of the MLP; all are drop-in means,
all enter the ensemble of Section~\ref{sec:diversity}, and their pairwise
error correlations are one of the measurements of this paper. We additionally
implement an encoder--decoder transformer
\citep{vaswani2017attention,dosovitskiy2021image} that tokenizes the 41 load
samples and decodes the field by cross-attention at the grid points; it is a
natural fourth architecture family, but we were unable to train it to the
level of the others on the hardware available for this study and report the
ensemble without it (Section~\ref{sec:diversity} returns to this).

All means are trained with AdamW and a cosine schedule, batches of 128--256,
for up to a few hundred epochs (Appendix~\ref{app:impl} lists every
hyperparameter).
During training each batch is reflected with probability $1/2$ (load reversed,
target field reflected along the first grid coordinate); at test time each
model is evaluated as $\frac12(f(u)+T f(Su))$. Proposition~\ref{prop:sym}
guarantees the test-time step cannot hurt on average, and the ablation
confirms small consistent gains from both steps. Optionally, the training
loss carries a kernel-flow term $\beta\,e_2$ computed from the batch
(Section~\ref{sec:theory-kf}) with a Gaussian kernel on pooled penultimate
features whose log-bandwidth is trained jointly; this variant appears in the
ablation.

\subsection{Stacking}
With $M$ trained models $f_1,\dots,f_M$ we form the convex combination
$m(u)=\sum_m w_m f_m(u)$, $w\in\Delta^{M-1}$, minimizing the validation
relative error; the weights are initialized at the simplex minimizer of the
measured second-moment matrix of Proposition~\ref{prop:secmom} and polished
by a short search on the 1000 held-out samples (250 in the low-data
protocol). A per-pixel variant allows the weights (plus an intercept) to
vary over the grid, ridge-fitted on half the validation split and accepted
only when it beats the global weights on the other half; it is the variant
the final pipeline uses. Stacking on a held-out split rather than uniform
averaging costs nothing, guards against a weak member
\citep{wolpert1992stacked}, and is statistically almost free at this size
(Proposition~\ref{prop:stack}).

\subsection{Kernel corrections}
\label{sec:method-kernel}
The stacked mean is then corrected by kernel ridge regression of its
residuals. Let $X=(u_1,\dots,u_n)$ be the training loads (standardized
coordinatewise), $R\in\mathbb R^{n\times1681}$ the matrix of training
residuals $v_i-m(u_i)$, and $k$ a Mat\'ern-$5/2$ kernel
$k(u,u')=\kappa\big(\|u-u'\|/s\big)$. The correction is
\[
\widehat r(u)=k(u,X)\,\alpha,\qquad \alpha=(K+n\lambda I)^{-1}R ,
\]
with $(s,\lambda)$ chosen on the validation split from a small grid ($s$ as a
multiple of the median pairwise distance, $\lambda\in\{10^{-7},10^{-5},10^{-3}\}$;
the grid is searched on an 8000-sample subsample and the chosen pair is refit
on all of $X$). The corrected surrogate is $m+\widehat r$. A second corrective
stage repeats the construction with the kernel evaluated on deep features,
the concatenated penultimate activations of the ensemble members
(reflection-averaged, standardized), fitted to the residuals of $m+\widehat
r$; it contributes a smaller improvement and is included when it helps on
validation. At $n=19000$ the exact solve is a single Cholesky factorization
of a $19000\times19000$ matrix, under half a minute in double precision on a
laptop CPU (measured in Section~\ref{sec:cost}), and prediction is two matrix
products; no inducing points, preconditioners, or stochastic solvers are
involved. The Gaussian process
reading of the same formulas supplies the posterior standard deviation
$P_\lambda(u)$ of \eqref{eq:power} at negligible extra cost (one triangular
solve per evaluation batch), which is the uncertainty estimate studied in
Section~\ref{sec:experiments} and certified by Theorem~\ref{thm:or}.

Two remarks on scope. Nothing in the construction uses properties of this
particular PDE beyond the verified reflection symmetry, so the recipe
(native input parametrization, symmetrized accurate means, validated kernel
correction of the residual, posterior standard deviation as certificate)
transfers to other operator learning problems with low-dimensional input
parametrizations. And the pipeline degrades gracefully: dropping the feature
stage, the stacking, or the corrections recovers progressively simpler
methods whose individual numbers appear in the ablation table.

\section{Results on the structural-mechanics benchmark}
\label{sec:experiments}
% Numbers inserted from runs/*.json at freeze; macros in main.tex hold the
% reported values.

\begin{figure}[t]
\centering
\includegraphics[width=\linewidth]{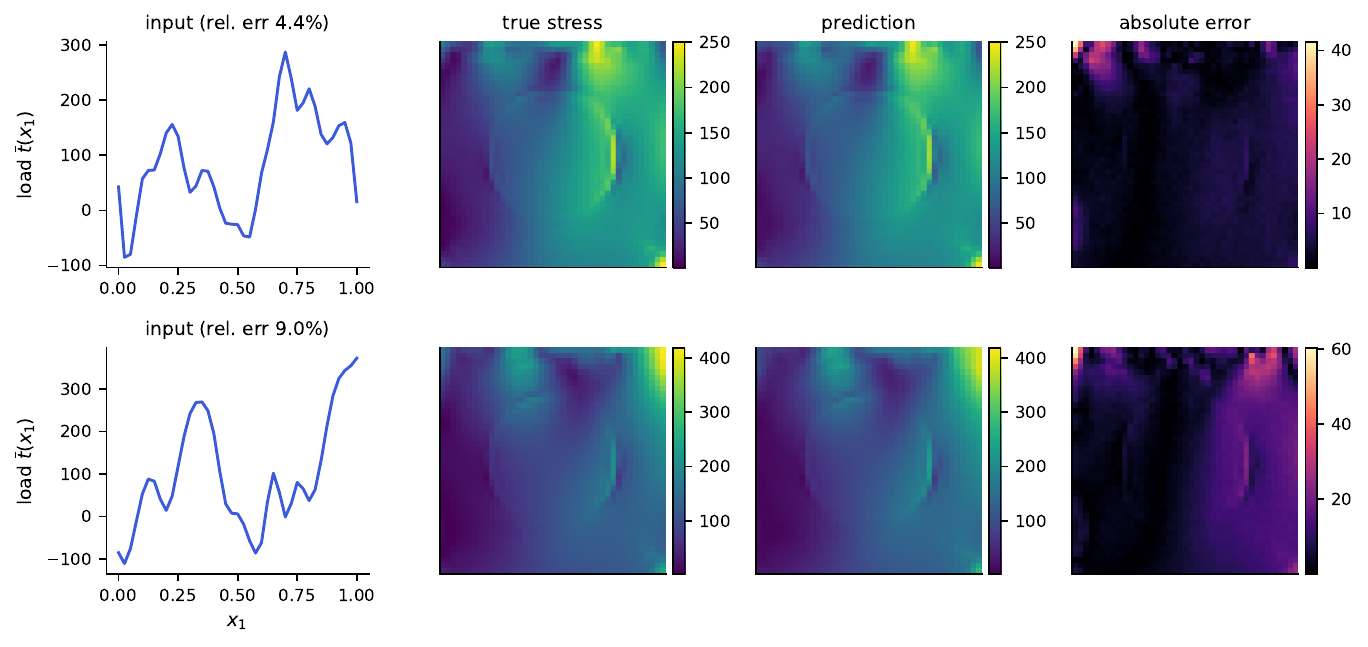}
\caption{Example predictions of the full pipeline. Top: a median-error test
case; bottom: a case at the 98th error percentile. Columns: input load, true
von Mises stress, prediction, and pointwise absolute error (note the smaller
scale). Errors concentrate in the high-traction corner, as also reported by
\citet{mora2025operator}.}
\label{fig:fields}
\end{figure}

\subsection{Main comparison}
\label{sec:main-results}

Table~\ref{tab:main} reports the high-data protocol. The published numbers
are quoted from \citet{dehoop2022cost} (as tabulated by
\citealt{batlle2024kernel}) and from \citet{batlle2024kernel}; our rows are
computed under the split of Section~\ref{sec:benchmark}, which gives our
methods strictly no more training information than the baselines had. The
kernel ridge row reproduces the published kernel result almost exactly
(\errKRRload\% against 5.18\%), which we take as evidence that the protocols
are aligned; the gap opened by the rest of the pipeline is therefore not an
artifact of evaluation choices. The full pipeline reaches \errBest\%. This is
below every published method other than PARA-Net (4.55\%), which it matches at
the reported two-decimal precision; since our own stacking configurations vary
by a hundredth or two among themselves (the ablation below moves between
4.65\% and 4.67\% under changes that should not matter), we read the two as
tied and do not claim to separate them. The margin over the FNO (4.76\%), PCA-Net
(4.67\%), DeepONet (5.20\%) and the optimal-recovery kernel (5.18\%) is
several times that spread. Section~\ref{sec:diversity} argues the tie is better read as a floor the
data impose than as a coincidence of tuning.

\begin{table}[t]
\centering
\caption{Structural mechanics, high-data protocol (20000 training samples,
20000 test). Mean relative $L^2$ test error. Rows above the rule are
published results on the same data; rows below are ours (mean over the fixed
split; TTA denotes reflection averaging at test time).}
\label{tab:main}
\begin{tabular}{lcc}
\toprule
Method & Error (\%) & Parameters \\
\midrule
DeepONet \citep{dehoop2022cost} & 5.20 & -- \\
FNO \citep{dehoop2022cost} & 4.76 & -- \\
PCA-Net \citep{dehoop2022cost} & 4.67 & -- \\
PARA-Net \citep{dehoop2022cost} & 4.55 & -- \\
Optimal-recovery kernel \citep{batlle2024kernel} & 5.18 & -- \\
\midrule
Kernel ridge on loads (ours) & \errKRRload & -- \\
Residual MLP + reflection TTA & \errMLP & 4.9M \\
\ \ + residual kernel correction (full pipeline) & \errBest & 4.9M \\
\bottomrule
\end{tabular}
\end{table}

\begin{table}[t]
\centering
\caption{Low-data protocol (1250 training samples, 20000 test). Published
rows from \citet{mora2025operator}. Our pipeline sees only the 1250 labels
(250 of them used as the validation split).}
\label{tab:lowdata}
\begin{tabular}{lc}
\toprule
Method & Error (\%) \\
\midrule
DeepONet & 8.70 \\
GP, optimal recovery \citep{batlle2024kernel} & 6.95 \\
Zero-mean GP \citep{mora2025operator} & 6.74 \\
FNO & 6.62 \\
FNO-mean GP \citep{mora2025operator} & 6.49 \\
\midrule
Kernel ridge on loads (ours, multiscale) & \errKRRloadLow \\
Full pipeline (ours) & \errBestLow \\
\bottomrule
\end{tabular}
\end{table}

\subsection{Ablation}
\label{sec:ablation}
Table~\ref{tab:ablation} builds the pipeline up one component at a time under
the high-data protocol. The kernel ridge on loads and the reflection-averaged
neural mean start at \errKRRload\% and \errMLP\% respectively. Stacking is
trivial with a single high-data mean and leaves it unchanged; the value of the
combination appears in the correction stage, where regressing the neural
mean's residual with the validated Mat\'ern kernel lowers the error to
\errBest\%. The feature-space and second input-space corrections did not
improve validation error beyond the first correction with a single mean and so
were not selected; we expect them to contribute more with a larger and more
diverse ensemble, and we report the stages that the validation split actually
chose. Two smaller effects are worth isolating. Reflection test-time averaging
lowers the MLP test error by \ttaGainMLP{} points, consistent with
Proposition~\ref{prop:sym}, which guarantees the direction of the change.
Adding the kernel-flow regularizer (Section~\ref{sec:theory-kf}) to the
low-data MLP did not help here (it moved test error from 5.64\% to 5.87\%);
Proposition~\ref{prop:kf} says what the term estimates, not that estimating it
is useful on every problem, and on this smooth-output benchmark the plain
metric loss was already a good proxy. Nor does a richer correction kernel help:
replacing the single tuned Mat\'ern with a sum of Mat\'ern kernels at several
bandwidths leaves the validation error unchanged to two decimals, which is
consistent with the residual being close to kernel-irreducible in the input
geometry, and is why the boosted correction stages of
Section~\ref{sec:method-kernel} were not selected. One positive lesson does
emerge from the refiner variants. Conditioning the refiner on the kernel
method's prediction (test error 4.73\%) beats conditioning it on another
network's prediction (4.84\%), even though the two conditioning fields have
almost the same accuracy; the kernel field carries information complementary to
the network's inductive bias, whereas a second network's field is largely
redundant. Enlarging the ensemble beyond the residual family is the subject of
the next subsection, where the returns of each addition turn out to be
predictable in advance from the correlation structure of the members' errors.

\begin{table}[t]
\centering
\caption{Building the high-data pipeline. Mean relative $L^2$ test error under
the 20000-sample protocol; ``+'' rows are cumulative.}
\label{tab:ablation}
\begin{tabular}{lc}
\toprule
Stage & Test error (\%) \\
\midrule
Kernel ridge on loads & \errKRRload \\
Reflection-averaged neural mean & \errMLP \\
\ \ + kernel-conditioned refiner, stack & \errStackFam \\
\ \ + diverse members (Section~\ref{sec:diversity}), per-pixel stack & \errStack \\
\ \ + residual kernel correction & \errBest \\
\bottomrule
\end{tabular}
\end{table}

\subsection{Architectural diversity and the shared residual}
\label{sec:diversity}

\begin{figure}[t]
\centering
\includegraphics[width=0.62\linewidth]{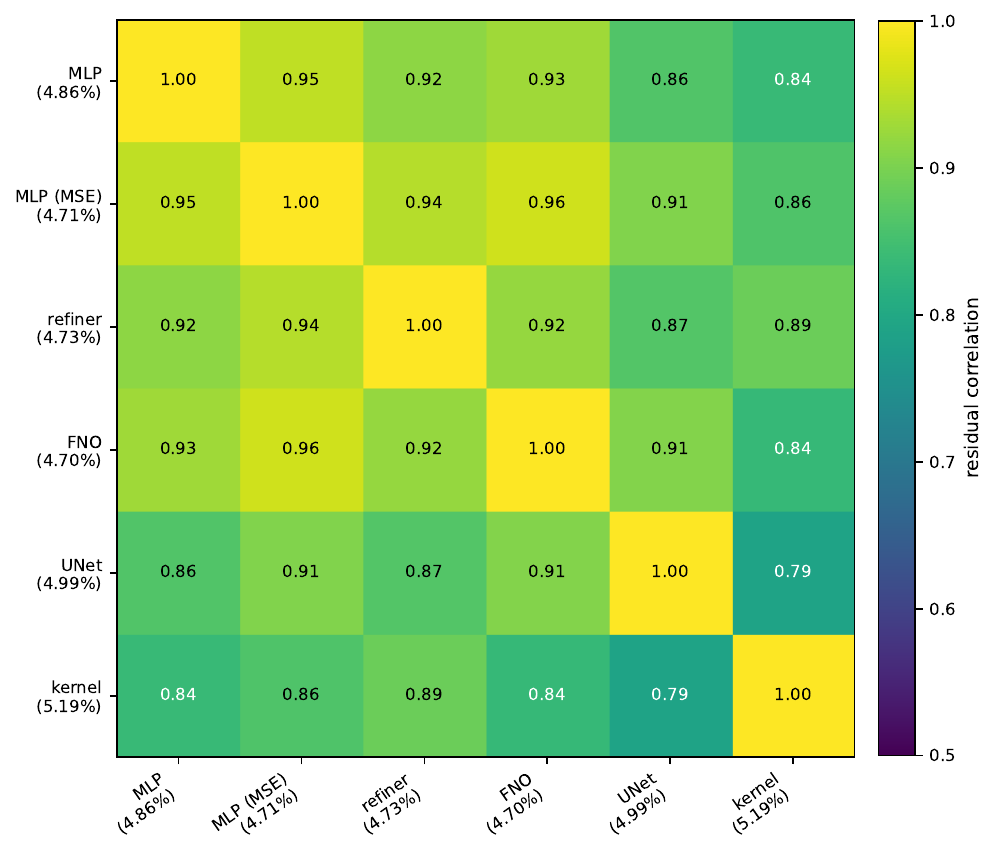}
\caption{Correlation of per-sample normalized residuals between ensemble
members on the validation split (test-set values agree to two decimals).
Every pair of trained networks correlates between \corrNetLo{} and
\corrNetHi{}, the convolutional UNet being the least like the rest; the kernel
predictor, the most alien member by construction, still correlates
\corrKerLo{}--\corrKerHi{} with all of them.}
\label{fig:corr}
\end{figure}

The natural reading of Table~\ref{tab:main} is that the remaining error is a
modeling problem: different architectures make different mistakes, so a more
diverse ensemble should stack to a lower number. We tested this directly. To
the residual family of Section~\ref{sec:method} we added a Fourier neural
operator, a UNet, and a variant of the MLP trained on normalized mean squared
error instead of the metric, each trained to the published single-model level
or better (Table~\ref{tab:members}). The test refutes the reading.
Figure~\ref{fig:corr} shows the correlation of per-sample normalized residuals
between members: every pair of trained networks, across three architecture
families and two losses, correlates between \corrNetLo{} and \corrNetHi{}, and
the kernel predictor, the most alien member by construction, still correlates
\corrKerLo{}--\corrKerHi{} with all of them. The members are not making
different mistakes. They are making the same mistake with
small private variations.

The second-moment identity of Proposition~\ref{prop:secmom} makes the
consequence quantitative before any weights are fitted. The matrix $S$
measured on the validation split predicts both the optimal convex weights and
the achievable stack risk: for the six-member ensemble the predicted
root-mean-square relative error of the best mixture is $4.97\%$, the fitted
stack realizes a mean of $4.61\%$, and their ratio is the dispersion factor
$\approx0.93$ that is stable across every pipeline we ran. The weights the
prediction assigns from $S$ alone, without ever evaluating the metric,
concentrated on the refiner and the FNO with a few tenths each on the UNet and
the MSE-trained network, a few percent on the kernel, and zero on the plain
MLP whose error is already spanned by the better members of its family, match
the weights the direct metric search finds, as part~(i) of the proposition
predicts from the measured $(e_1,e_2,\varrho)$. Stacking does exactly what the
correlations permit, and members like these permit little: part (ii) puts the
infinite-ensemble floor of an equicorrelated family at $e\sqrt\varrho$, which
at $e\approx4.7\%$ and a mean pairwise $\varrho\approx0.9$ is about $4.5\%$.
Allowing the
weights to vary over the grid (an affine per-pixel stack, fitted by ridge
regression on half the validation split and accepted only because it beat the
global weights on the other half) recovers a little of what global weights
cannot see, and the kernel correction adds a few hundredths on top; that is
the \errBest\% of Table~\ref{tab:main}, and it is consistent with the floor
just computed.

\begin{figure}[t]
\centering
\includegraphics[width=\linewidth]{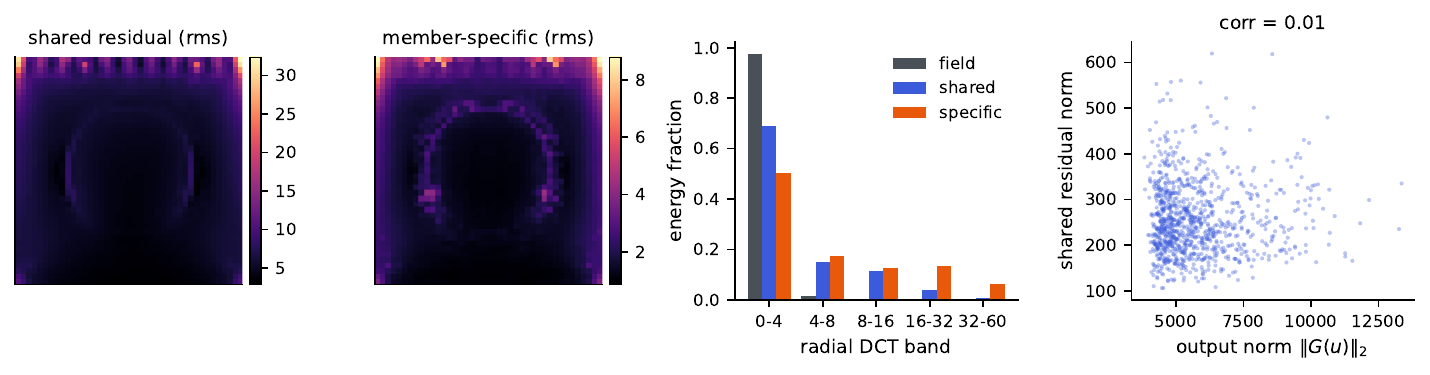}
\caption{Anatomy of the shared residual $c$ (across-member mean of the
residual fields; refiner, FNO and kernel members). Left to right: rms of $c$
over the grid; rms of the member-specific remainders; radial DCT band
energies of the stress fields, of $c$, and of the specific parts; and the
norm of $c$ against the output norm, whose correlation is $0.01$.}
\label{fig:shared}
\end{figure}

Where does the common mistake live? Write $r_m$ for member $m$'s residual
field on a validation case and $c=\frac1M\sum_m r_m$ for the across-member
mean, the component that no amount of uniform averaging removes. Measured
over the validation split and three families (refiner, FNO, kernel), the
shared component carries $90\%$ of the average residual energy
(Figure~\ref{fig:shared}). Three of its properties matter. Spatially, its
energy concentrates at the two corners of the loaded edge, where the traction
boundary condition meets the lateral supports; relative to the local field
scale it is three to four times larger there than the grid average, and these
are precisely the locations where the finite element solution is least
accurate and where interpolation to the $41\times41$ grid is most strained.
Spectrally, it is enriched in high radial frequencies relative to the stress
fields themselves: the fields put $97.5\%$ of their energy below radial
wavenumber 4, the shared residual only $69\%$. And its magnitude is
statistically independent of everything we can compute from the input: the
correlation of $\|c\|$ with the output norm is $0.01$, with the load norm
$0.01$, with the load's total variation $0.01$. A component that no
architecture avoids, that no input statistic predicts, that lives at the
stress concentrations, and that fixed-scale additive noise would reproduce is
most economically read as noise of the data-generating process itself, finite
element and grid-interpolation error at the singular corners, not as
approximation error of the surrogates.

\begin{figure}[t]
\centering
\includegraphics[width=\linewidth]{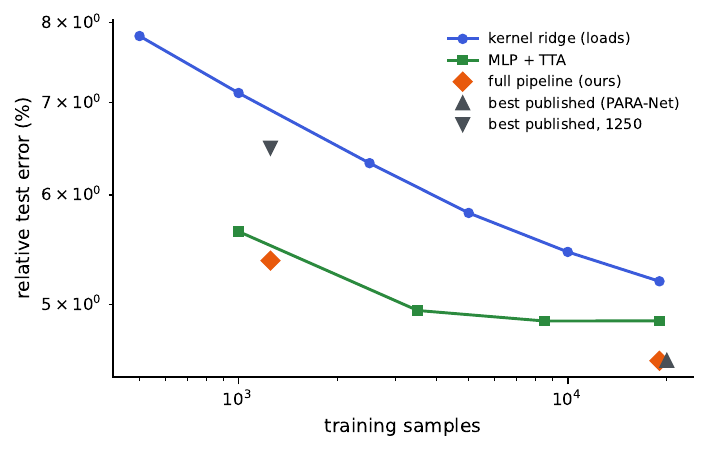}
\caption{Relative test error against training set size for the kernel ridge on
loads (log-log), with the full pipeline and the best published results marked
in both regimes.}
\label{fig:scaling}
\end{figure}

Two independent measurements corroborate this reading. First, the residual
kernel correction, which regresses the stack's residual on the load, improves
the stack by only a few hundredths of a point at $n=19000$ whatever the
kernel scale: at this sample size the shared residual is not a function of
the load in any way a Mat\'ern RKHS on $\mathbb R^{41}$ can see. Second, the
error stops responding to data. Under identical recipes, the MLP's test error
is $4.95\%$ with $3500$ training samples, $4.86\%$ with $8500$, and $4.86\%$
with $19000$ (Figure~\ref{fig:scaling}); the kernel ridge curve still falls,
at roughly $n^{-0.11}$, but toward the same region. A modeling limitation
should yield to capacity, diversity, or data. This error yields to none of
them.

We draw two conclusions. First, the published plateau of
Table~\ref{tab:main}, five methods within two thirds of a point of each other
after years of architectural progress, is not evidence that some sixth
architecture is missing; every measurement above is what a benchmark whose
achievable error is set by its data would produce. This is an inference from
surrogate-side measurements --- certifying it would take regenerated data,
mesh-refinement checks, or repeated solver evaluations, which we did not
perform --- but on this reading, numbers near $4.5\%$ sit within a few
hundredths of the limit, which is where our pipeline lands (\errBest\%), and
material further progress on this benchmark would come from regenerating the
data (finer meshes near the corners, higher-order elements, output grids that
resolve the concentrations) rather than from new surrogates. Second,
the practical recipe survives the reinterpretation with its economics
clarified: symmetrized accurate means capture what is learnable, stacking
buys exactly the decorrelation the members possess (little, here), and the
kernel correction certifies and mops up the load-visible remainder. The
components are worth their cost in that order.

\begin{table}[t]
\centering
\caption{Ensemble members, high-data protocol. Mean relative $L^2$ test error
with reflection averaging; every member is selected on the validation split.}
\label{tab:members}
\begin{tabular}{llc}
\toprule
Member & Family / loss & Error (\%) \\
\midrule
Residual MLP & MLP, metric loss & \errMLP \\
Residual MLP & MLP, normalized MSE & \errMSE \\
Kernel-conditioned refiner & MLP on $(u,\mathrm{KRR}(u))$ & \errRef \\
FNO & spectral, metric loss & \errFNOG \\
UNet & convolutional, metric loss & \errUNetG \\
Kernel ridge on loads & Mat\'ern-$5/2$ & \errKRRload \\
\midrule
Per-pixel stack (validation-fitted) & -- & \errStack \\
\ \ + residual kernel correction & -- & \errBest \\
\bottomrule
\end{tabular}
\end{table}

\subsection{Uncertainty quantification}
\label{sec:uq}
\begin{figure}[t]
\centering
\includegraphics[width=\linewidth]{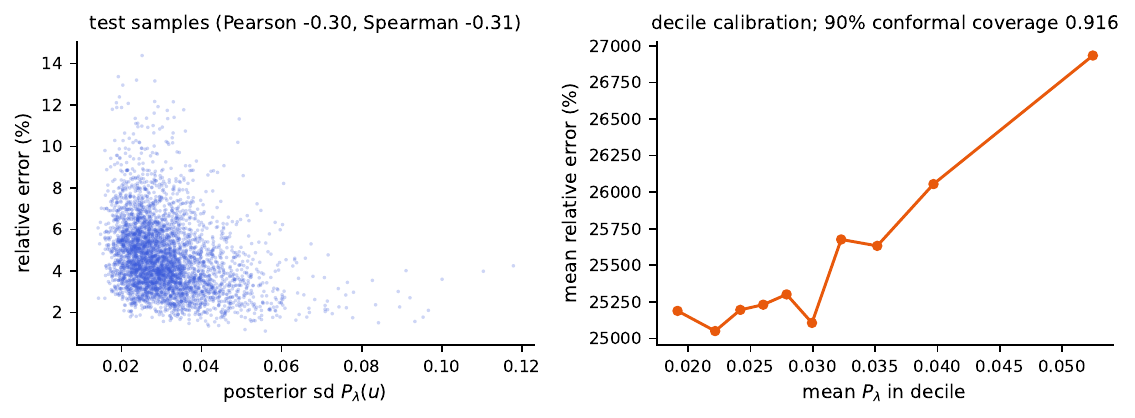}
\caption{Left: the Gaussian process posterior standard deviation $P_\lambda$
against the corrected surrogate's relative error on the test set; the
association is weak and, because of the output-amplitude confound, slightly
negative. Right: decile calibration of $P_\lambda$ against absolute error, the
theory-consistent quantity. A distribution-free conformal rescaling of
$P_\lambda$ attains \uqCover\% coverage at the nominal \uqCoverNom\% level.}
\label{fig:uq}
\end{figure}

Theorem~\ref{thm:or} certifies the corrected surrogate's error by
$\|G-m\|_K\,P_\lambda(u)$, and it is tempting to read the design factor
$P_\lambda$ as a pointwise error bar. On this benchmark that reading fails, in
an instructive way. Table~\ref{tab:norms} confirms the operator factor behaves
as the theory predicts: the computable squared RKHS norm of the fitted
correction is about \normRatio$\times$ smaller when the kernel regresses
ensemble residuals than when it regresses raw stress fields, so an accurate
mean genuinely leaves a smoother remainder and a tighter certificate (the
bound, linear in the norm, by the square root of that). But the design factor
$P_\lambda$ correlates only weakly with the corrected surrogate's absolute
error (Pearson \uqCorrAbs), and \emph{negatively} with the benchmark's
relative error. The cause is measurable: $P_\lambda$ is large for loads far
from the training set, those loads tend to have large amplitude, large-amplitude
loads produce large-norm stress fields, and dividing by the output norm makes
their relative error small. Indeed $P_\lambda$ correlates \uqCorrNorm{} with
the output norm (Figure~\ref{fig:uq}). The error is dominated by the neural
mean, whose mistakes are not a function of the input-space geometry that
$P_\lambda$ sees, so a purely input-space power function cannot rank them.

The obvious alternative fares no better, and with the ensemble of
Section~\ref{sec:diversity} the question can be answered rather than left
open. Ensemble disagreement, the spread of the members' predictions, is the
standard deep-ensemble error signal; on the full cross-architecture ensemble
its correlation with the corrected surrogate's absolute error is
\uqDisAbsDiv, no better than the \uqDisAbs{} of the like-architecture
ensemble. Section~\ref{sec:diversity} says why it cannot do better here:
disagreement measures the member-specific components of the error, which
carry under a tenth of the residual energy, while the ranking signal for the
shared component is invisible to any within-ensemble statistic, because the
members agree precisely where they are jointly wrong.

What does hold is coverage, for free. A distribution-free conformal rescaling,
taking the 0.9-quantile of $\|e\|_2/P_\lambda$ on the validation split and
scaling $P_\lambda$ by it, yields a band with \uqCover\% empirical coverage on
the test set at the nominal \uqCoverNom\% level. Conformal validity needs no
correlation between the score and the error, only exchangeability, so it
survives the confound that defeats the ranking. One honesty note, expanded in
Section~\ref{sec:theory-conf}: the same validation split served model
selection, so the exchangeability behind the exact guarantee is approximate
here, and the observed coverage should be read as empirical; a calibration
split untouched by selection would make the guarantee exact. The lesson is narrow but real:
a valid pointwise bound (Theorem~\ref{thm:or}) is not automatically a useful
pointwise indicator, the standard uncertainty signals can both fail on the
same problem, and a relative-error metric has to have its normalization
accounted for before a posterior standard deviation means what it appears to.

\begin{table}[t]
\centering
\caption{Squared RKHS norm of the fitted kernel interpolant,
$\operatorname{tr}(\alpha^\top K\alpha)$, when the same validated kernel and
design regress raw stress fields versus ensemble residuals.}
\label{tab:norms}
\begin{tabular}{lc}
\toprule
Kernel regresses & $\|\widehat r\|_K^2$ \\
\midrule
Raw stress fields ($m=0$) & $1.39\times10^{10}$ \\
Ensemble residuals & $5.14\times10^{7}$ \\
\midrule
Ratio & \normRatio$\times$ \\
\bottomrule
\end{tabular}
\end{table}

\subsection{Spectra and effective dimension}
\label{sec:spectra}
\begin{figure}[t]
\centering
\includegraphics[width=\linewidth]{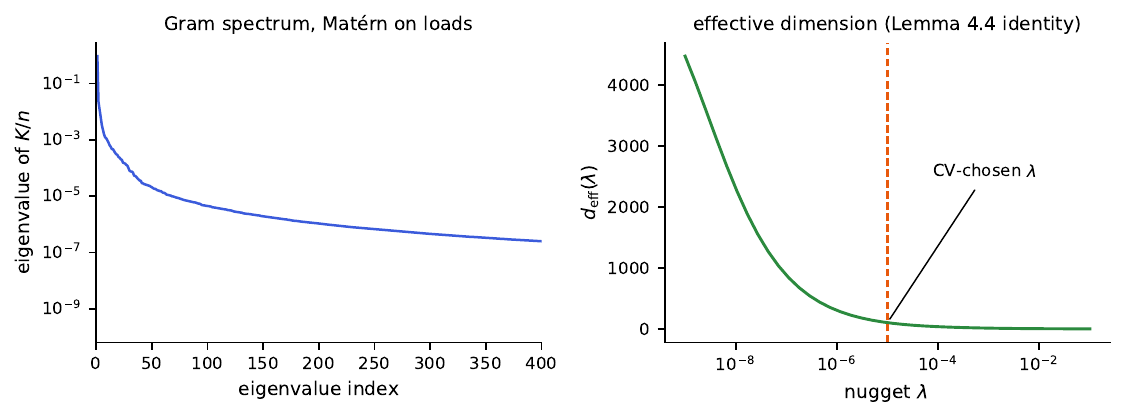}
\caption{Left: eigenvalues of the Mat\'ern Gram matrix $K/n$ on the loads
(log scale); the spectrum decays by many orders of magnitude within a few
hundred modes. Right: the effective dimension $d_{\mathrm{eff}}(\lambda)$ from
the exact identity of Lemma~\ref{lem:deff}, with the cross-validated nugget
marked.}
\label{fig:spectra}
\end{figure}

Figure~\ref{fig:spectra} shows the spectrum of the Mat\'ern Gram matrix on the
loads and the effective dimension $d_{\mathrm{eff}}(\lambda)$ computed from it
by the exact identity of Lemma~\ref{lem:deff}. The spectrum decays quickly: a
few hundred eigenvalues carry essentially all the mass, the tail falling many
orders of magnitude below the leading modes. At the cross-validated nugget the
effective dimension is a few hundred out of $n=19000$. This is a statistical
statement, not a computational one --- the dense factorization costs its full
$n^3$ operations regardless --- but it is why the correction generalizes
despite interpolating nineteen thousand points: the problem the kernel solves
has the statistical dimension of the retained spectrum, not of the sample. The
cross-validated $\lambda$ sits at the shoulder of the $d_{\mathrm{eff}}$ curve,
past the leading modes and into the rapidly decaying tail, matching the reading
of Lemma~\ref{lem:deff}.

\subsection{Cost and reproducibility}
\label{sec:cost}
The pipeline was built and run on a single laptop. The kernel stages are exact:
one Cholesky factorization of the $19000\times19000$ Mat\'ern Gram matrix in
double precision takes $24$ seconds on the laptop's CPU ($16$ threads,
OpenBLAS; the matrix itself is $2.9$\,GB), and prediction is two matrix
products; there are no inducing points, stochastic solvers, or
preconditioners, and no GPU is needed for the correction. The factorization
performs the full $n^3/3\approx2.3\times10^{12}$ floating-point operations ---
numerical low rank does not reduce the work of a dense Cholesky --- and the
measured rate, about $100$ gigaflops per second, is simply what a current
multicore CPU delivers; exactness at this $n$ costs half a minute, not a
cluster. What the low effective dimension of Section~\ref{sec:spectra} buys is
statistical rather than computational: it describes where the solve's
information concentrates at the working nugget, not its flop count. The neural
means are small by the standards of the literature (a few million parameters)
and train in the metric itself. All splits
are fixed by a published permutation seed, the input is consumed as the
41-dimensional load after the exact broadcast check of
Section~\ref{sec:benchmark}, and every model is selected on the validation
split and evaluated once on the test set; we release the splits, the trained
means, and the code so the table can be regenerated end to end.

\section{The OCO-2 radiative-transfer emulator}
\label{sec:oco2}
The structural-mechanics benchmark put the neural mean and the kernel in the
balanced regime. To see the other regime we apply the same components to the
radiative-transfer emulation problem of \citet{susiluoto2025radiative}: per
spectral band of the OCO-2 instrument, learn the map from a reduced
atmospheric state ($20$--$24$ dimensions after the dimension reduction of
that paper) to the reduced radiance ($40$ PCA coefficients of the
monochromatic spectrum). The OSF project (\texttt{u2t8a}) publishes a training
pool of $20000$ pairs and, separately, $2000$ test states together with the
kernel-flow emulator's own predictions on them; we verified the two sets share
no point. We split the pool $18000/2000$ into training and validation, and
every choice in our pipeline --- the network checkpoints, the kernel head's
length scale and nugget, and the per-coordinate winners of the combination
below --- is made on that validation split; the $2000$ public test states are
used once, for the final numbers. The comparison is therefore computed on
identical test points against the published emulator itself rather than
against our reimplementation of it.

Two error metrics matter, and they disagree in an instructive way. The
\emph{reduced} metric is the relative $L^2$ error on the $40$ standardized
coefficients. The \emph{radiance} metric maps predictions back to the
monochromatic spectrum through the stored PCA projection and norms before
comparing; because the projection is orthogonal, the error \emph{numerator} on
the reduced side is exactly the diagonally weighted norm
$\|s_z\odot(\hat z-z)\|$, whose weights concentrate almost entirely on the
first few coefficients (the denominator is the full reconstructed-radiance
norm, which adds the reconstruction offsets). It is this diagonal weighting of
the residual that the per-coordinate combination exploits.

\begin{table}[t]
\centering
\caption{OCO-2 O2 band, scored on the $2000$ public test states, which are
disjoint from the training pool; all model selection happened on a validation
split of the pool. The
kernel-flow row is the emulator of \citet{susiluoto2025radiative}, scored
from its own stored predictions. ``Features'' means the penultimate
activations of the trained network; the kernel head is the same exact
Mat\'ern solve as everywhere else in this paper. The two raw-input rows are the
diagnostic-script fit (a $6000$-point solve at a single median length scale),
lighter than the feature head's full-grid fit; the anisotropy experiment of
Section~\ref{sec:theory-aniso}, which tunes the raw kernel fully, confirms it
stays an order of magnitude above the feature head, so the gap is not an
artifact of this.}
\label{tab:oco2}
\begin{tabular}{lcc}
\toprule
model & reduced & radiance \\
\midrule
Mat\'ern kernel, raw input, one length scale & 40.47\% & 0.240\% \\
Mat\'ern kernel, raw input, sensitivity-scaled (ARD) & 30.01\% & --- \\
kernel-flow emulator \citep{susiluoto2025radiative} & 16.89\% & 0.0448\% \\
residual MLP, flat metric & 3.99\% & 0.159\% \\
residual MLP, radiance metric & 46.3\% & 0.0291\% \\
Mat\'ern head on the flat network's features & 3.82\% & 0.0758\% \\
Mat\'ern head on the weighted network's features & 20.0\% & 0.0282\% \\
per-coordinate combination of the above & \textbf{3.83\%} & \textbf{0.0267\%} \\
\bottomrule
\end{tabular}
\end{table}

\begin{figure}[t]
\centering
\includegraphics[width=\linewidth]{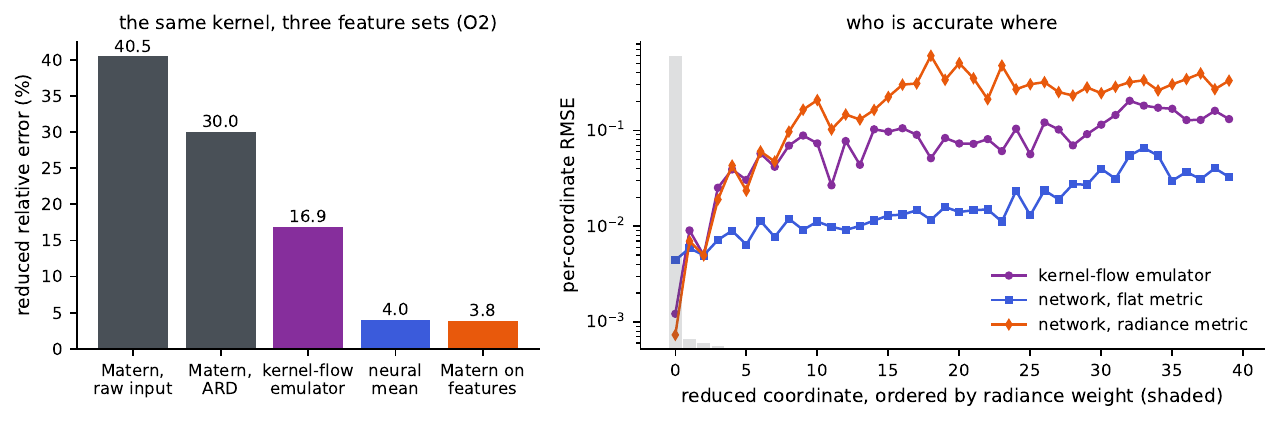}
\caption{Left: the representation ladder on the O2 band. The same exact
Mat\'ern machinery moves from $40\%$ on the raw input to $3.8\%$ on the
trained network's features; the kernel-flow emulator and the network sit in
between. Right: per-coordinate test RMSE with the reduced coordinates ordered
by their radiance weight (shaded bars). The per-output-tuned kernel-flow
emulator is most accurate exactly on the heavily weighted coordinates, the
flat-metric network on the many light ones, and the radiance-trained network
closes the gap where the weight lives; the per-coordinate combination takes
each coordinate from whichever model wins it.}
\label{fig:oco2}
\end{figure}

Table~\ref{tab:oco2} carries four findings, and Figure~\ref{fig:oco2} shows
the two central ones. First, the representation
ladder: the same exact kernel machinery scores $40\%$ on the raw input,
$30\%$ after rescaling each input coordinate by its measured sensitivity (the
metric term of Proposition~\ref{prop:aniso}: the effective rank here is
$17$ of $20$, so the proposition predicts a constant-factor gain and no
change of rate, the $40\%\!\to\!30\%$ the ARD row records), and
$3.82\%$ on the trained network's features, past the network's own head. The
raw-input kernel was limited by its features, not by anything kernel-shaped,
and the deep kernel head, an exact solve on learned features, is the
strongest single model on the reduced metric --- though its margin over the
network it feeds on ($3.82\%$ against $3.99\%$, one seed each) is small, and
we lean on the ladder's order of magnitude, not on that gap. Learning the metric rather than
setting it by hand does not change this reading. A per-dimension metric fit by
the kernel-flows cross-validation loss of \citet{owhadi2019kernel} reaches the
same $30\%$ on the raw state, and a low-rank Mahalanobis metric with a thousand
more parameters does not improve on it; on the features, where the network has
already conditioned the representation, a learned metric gives no gain over the
isotropic kernel. This is what Proposition~\ref{prop:aniso} anticipates: the
metric is a constant-factor lever, and the representation is the
order-of-magnitude one.

Second, the training metric behaves as a budget. The flat-trained and
radiance-trained rows are one architecture and two losses, and each wins the
metric it was trained in by a factor of five or more while losing the other.
The mechanism is visible per coordinate: the radiance metric concentrates on
the leading coefficient, the kernel-flow emulator (whose lengthscales are
tuned per output) predicts that coefficient to $0.0012$ root mean square
against the flat network's $0.0054$, while the flat network is three to four
times more accurate on the trailing thirty coordinates that the radiance
barely sees. Training the network in the radiance metric closes exactly that
gap. A constant-elasticity model of this reallocation is too crude --- the
per-coordinate exponents scatter widely, and coordinate errors do not
decouple because the network shares capacity --- but the direction is robust,
and the practical rule is plain: train in the metric you report.

Third, the same-class floor of the structural-mechanics study reappears here,
one level down. The residuals of independently seeded copies of the flat
network correlate at $0.78$ on the O2 band and at $0.97$ and $0.96$ on WCO2
and SCO2, so part~(ii) of Proposition~\ref{prop:secmom} puts the
infinite-seed floors of the flat network at $3.9\%$, $16.2\%$ and $8.0\%$, and
its seed ensemble sits there. Passing below the floor took a different kind of
member: the Mat\'ern head on learned features reaches $3.82\%$, $16.1\%$ and
$7.96\%$, and the final combination $3.83\%$, $16.1\%$ and $7.96\%$. Ensembling
within an architecture class saturates by the same second-moment law on both
problems; what moved the OCO-2 numbers an order of magnitude was changing what
the members are (learned features under the kernel), not how many there are.

\begin{figure}[t]
\centering
\includegraphics[width=\linewidth]{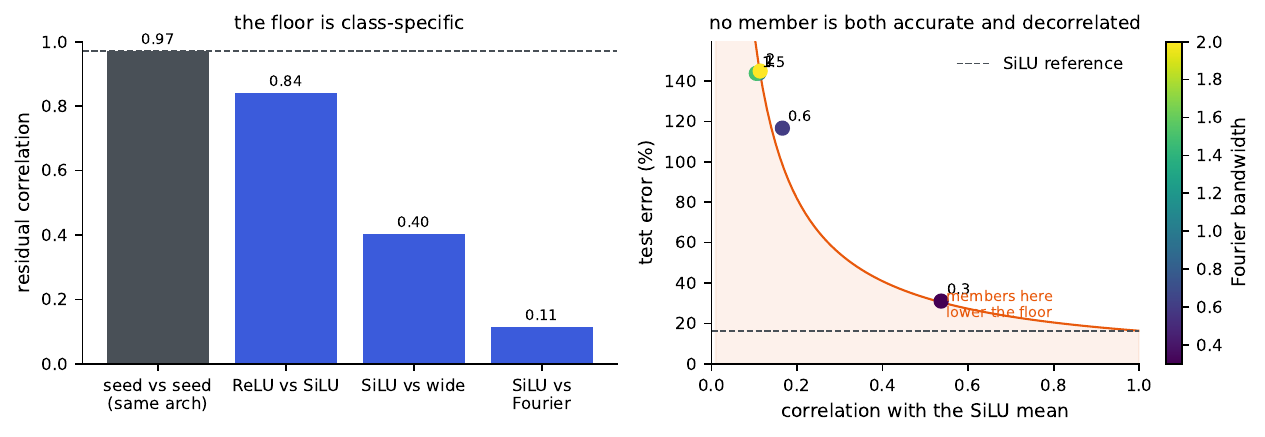}
\caption{The ensembling floor on WCO2. Left: residual correlations. Two seeds
of one architecture correlate near the value that sets the floor (dashed);
different architectures correlate far less, so the floor is a property of the
architecture class. Right: the Fourier-feature member across bandwidths. A
member helps the ensemble only in the shaded region, below the
$e_{\mathrm{ref}}/\varrho$ curve of Proposition~\ref{prop:secmom}(i); the
accurate settings are too correlated and the decorrelated settings too
inaccurate, so none lands there with room to spare.}
\label{fig:floor}
\end{figure}

The floor is a property of the architecture class, not of the problem, and it
can be probed directly. Different architectures decorrelate: on WCO2 two
residual MLPs of different activation correlate at $0.84$, a wide shallow
network against the residual MLP at $0.40$, and a random-Fourier-feature
network against it at $0.08$ to $0.17$, all far below the $0.96$ of two seeds.
Yet the floor did not move, because its other input is accuracy: by
part~(i) of Proposition~\ref{prop:secmom} a member of error $e_2$ helps a
reference of error $e_1$ only when their correlation falls below $e_1/e_2$,
and across a bandwidth sweep the Fourier network was either accurate and
correlated (error $31\%$, correlation $0.54$, just above the $0.53$ threshold)
or decorrelated and inaccurate (error above $140\%$). No architecture we tried
was both accurate and decorrelated enough to lower the floor, which is the
honest statement of why the residual MLP is the member of record: not that
diversity is impossible, but that on this map no more accurate diverse member
was found. Figure~\ref{fig:floor} shows both halves of this: the correlations
that make the floor class-specific, and the accuracy-decorrelation trade-off
that keeps it in place. Adding the Fourier members to the per-coordinate
combination changes the reported error by less than a hundredth of a point,
in either direction, on all three bands: the honesty-split blend gives them no
weight.

Fourth, the regimes of Section~\ref{sec:theory-secmom} are decided by these
numbers before any stack is fit. Against the flat network the raw-input
kernel has $\varrho=0.14$ and $e_n/e_k=0.10$ (the network's $3.99\%$ over the
kernel's $40.47\%$): the condition of
Proposition~\ref{prop:secmom} fails and the kernel is dropped, in contrast to
structural mechanics where the same test keeps it. Because the radiance
metric is diagonal in the reduced coordinates, the per-coordinate
combination of Proposition~\ref{prop:percoord} is optimal for both metrics
simultaneously. On O2 and SCO2 this yields a single surrogate that beats the
kernel-flow emulator on both metrics at once: O2 reaches $3.83\%$ against
$16.89\%$ reduced and $0.0267\%$ against $0.0448\%$ radiance, and SCO2 reaches
$7.96\%$ against $16.14\%$ reduced and $0.043\%$ against $0.115\%$ radiance.
WCO2 is a partial exception, stated plainly. We did not produce a single
combined WCO2 surrogate, and its two best heads split the metrics: the
flat-feature head reaches $16.1\%$ reduced, below the emulator's $24.1\%$, but
$0.101\%$ radiance, \emph{above} the emulator's $0.060\%$; the radiance-trained
head reaches $0.035\%$ radiance but at $48\%$ reduced. On WCO2, then, each
metric is won by a different head, and unlike the other two bands no single
model beats the emulator on both. The code releases the per-band numbers
alongside this paper.

\subsection{The error is limited by data, and yields to more of it}
\label{sec:scaling}
\begin{figure}[t]
\centering
\includegraphics[width=\linewidth]{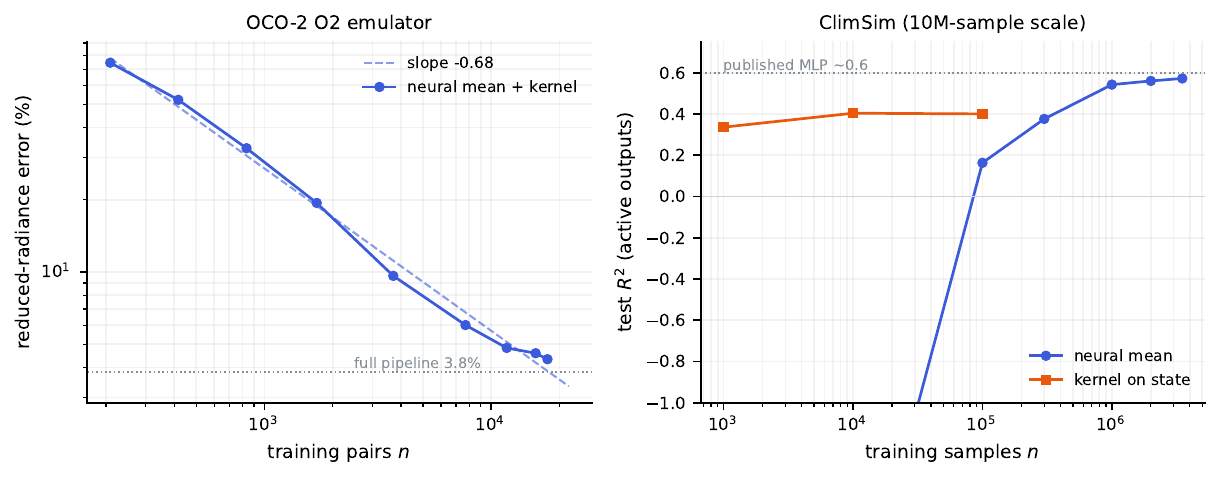}
\caption{Test error against training-set size, at fixed test set. Left: the
OCO-2 O2 task, reduced-radiance error of the corrected surrogate; it falls as a
power law (fitted slope $-0.68$) to $4.3\%$ at $n=17700$, approaching the
$3.8\%$ the full pipeline reaches on all $18000$ pairs (dotted). Right: ClimSim,
test $R^2$ over the outputs with non-negligible variance; the exact kernel is
flat and data-efficient while the neural mean, underfit at small $n$, overtakes
it near a few hundred thousand samples and closes on the published baseline
(dotted) by a few million.}
\label{fig:ocoscaling}
\end{figure}

The structural-mechanics error was flat in the sample size because it is set
by the noise of the finite element data; the emulator error is not, and the
contrast is the sharpest practical difference between the two problems. The
left panel of Figure~\ref{fig:ocoscaling} fixes the O2 task and varies only the
number of training pairs, scoring each on a disjoint held-out set. The
corrected surrogate's reduced-radiance error falls as a clean power law in $n$,
fitted slope $-0.68$, from $74\%$ at a few hundred pairs to $4.3\%$ at
$n=17700$, still descending at the largest sample size we can afford and
approaching the $3.8\%$ our full pipeline reaches on all $18000$ pairs
(Table~\ref{tab:oco2}). The exact kernel on the raw state improves far more
slowly, so at these sizes the network is the component that turns data into
accuracy. The residual on this task is sample-limited rather than
noise-limited: unlike the structural-mechanics floor it recedes as pairs are
added, so the largest lever on this problem is simply more of them, and pairs
are cheap because the forward model is a program. The complementary case is the
full-resolution $58\to3048$ map, where the error is instead flat in $n$ across
the few hundred retrievals we hold: that task is limited by its representation,
not its sample count, and the reduction the emulator applies is what moves it.

The same reading holds well beyond OCO-2, on a benchmark large enough to watch
the two regimes trade places. ClimSim \citep{yu2023climsim} is a climate
emulation dataset of about ten million samples that maps a $124$-dimensional
atmospheric state to $128$ physics tendencies, with published neural baselines.
The right panel of Figure~\ref{fig:ocoscaling} sweeps the training size across
more than three orders of magnitude at a fixed test set, in the coefficient of
determination the benchmark reports (over the outputs with non-negligible
variance). The exact kernel on the state is data-efficient: its $R^2$ is
already about $0.4$ at ten thousand samples and holds there. The neural mean is
data-hungry: badly underfit at a thousand samples, with $R^2$ well below zero,
it crosses the kernel near a few hundred thousand samples and keeps rising,
reaching $R^2=0.56$ at two million and $0.57$ at three and a half million,
closing on the published multilayer-perceptron baseline near $0.6$. Which member is the stronger one is
therefore a matter of the sample size alone---the kernel below the crossover,
the network above it---and the two regimes of
Section~\ref{sec:theory-secmom} gain a data axis: the accurate mean is worth
building precisely once there is enough data to train it past the kernel it
would otherwise defer to.

\section{Theory}
\label{sec:theory}
Throughout this section $\mathcal U=\mathbb R^p$ denotes the (discretized) input
space and $\mathcal V=\mathbb R^q$ the output space, $G:\mathcal U\to\mathcal V$
the target operator, $\mu$ the input distribution, and
$(u_1,v_1),\dots,(u_N,v_N)$ i.i.d.\ draws with $v_n=G(u_n)$; the data carry no
sampling noise in the usual sense, coming from a deterministic solver, though
Section~\ref{sec:diversity} has something to say about solver error. We write
$\ell(w,v)=\|w-v\|_2/\|v\|_2$ for the relative error and
$R(\widehat G)=\mathbb E_{u\sim\mu}\,\ell(\widehat G(u),G(u))$ for the risk, which is
the quantity reported in Section~\ref{sec:experiments}. The results below are
largely elementary, but each one licenses a specific design decision in the
pipeline of Section~\ref{sec:method}, and each has a measurable footprint in
the experiments: one statement per stage, from the symmetry handling and the
kernel-flow term through stacking, the residual correction, its coverage
guarantee, and the choice of nugget. Proofs are collected in
Appendix~\ref{app:proofs}.

\subsection{Symmetry averaging does not increase the risk}
\label{sec:theory-sym}

The elasticity problem behind the benchmark is invariant under the reflection
$x_1\mapsto 1-x_1$: the domain, the boundary partition, and the constitutive
model are all symmetric, the input distribution has a reflection-invariant
covariance, and reflecting the load reflects the stress field. On the discrete
grid this is expressed by two permutation matrices: $S\in\mathbb R^{p\times p}$
reverses the load samples and $T\in\mathbb R^{q\times q}$ reverses the rows of
the stress field. Both are orthogonal involutions. Section~\ref{sec:benchmark}
describes a direct data-driven check of the equivariance
$G(Su)=T\,G(u)$; we treat it as an assumption here.

\begin{proposition}[symmetrization]\label{prop:sym}
Assume $G(Su)=TG(u)$ for $\mu$-a.e.\ $u$, that $\mu$ is invariant under $S$,
and that $T$ is an orthogonal involution ($T^2=I$). For a measurable
$\widehat G:\mathcal U\to\mathcal V$ define its symmetrization
\[
\widehat G_S(u)\;=\;\tfrac12\left(\widehat G(u)+T\,\widehat G(Su)\right).
\]
Then, for every loss $\ell(\cdot,v)$ that is convex in its first argument and
satisfies $\ell(Tw,Tv)=\ell(w,v)$,
\[
\mathbb E_{u\sim\mu}\,\ell\big(\widehat G_S(u),G(u)\big)\;\le\;
\mathbb E_{u\sim\mu}\,\ell\big(\widehat G(u),G(u)\big).
\]
\end{proposition}

The relative error satisfies both hypotheses ($T$ orthogonal gives
$\ell(Tw,Tv)=\ell(w,v)$). Proposition~\ref{prop:sym} justifies the two uses of
the symmetry in the pipeline: averaging each trained model with its reflected
evaluation at test time can only help on average, and training on
reflection-augmented data is ordinary empirical risk minimization for the
symmetrized class. The measured effect is small but strictly positive for
every model we trained (Table~\ref{tab:ablation}).

\subsection{What the kernel-flow regularizer estimates}
\label{sec:theory-kf}

The kernel-flow (KF) loss of Owhadi and Yoo \citep{owhadi2019kernel}, in the
$\ell^2$ variant used by \citet{yoo2021deep} to regularize deep networks,
enters our ablation study as an optional term in the training of the neural
means. Its motivation is often stated informally (``a kernel is good if
half the data predicts the rest''). The following observation makes the
statement exact. Let $z_i=\phi_\theta(u_i)$ be the features of a batch
$B=\{1,\dots,b\}$ ($b$ even), let $k$ be a positive definite kernel on feature
space, and for $C\subset B$, $|C|=b/2$, let $I_C$ denote the $k$-interpolant
of $\{(z_j,v_j):j\in C\}$. The batch KF loss is
\begin{equation}\label{eq:kf}
e_2(\theta;C)\;=\;\sum_{i\in B}\big\|v_i-I_C(z_i)\big\|_2^2 .
\end{equation}

\begin{proposition}[KF loss is a leave-half-out estimate]\label{prop:kf}
Let $C$ be drawn uniformly among the subsets of $B$ of size $b/2$ and assume
$k$ restricted to $\{z_i\}_{i\in B}$ is strictly positive definite. Then
$e_2(\theta;C)=\sum_{i\in B\setminus C}\|v_i-I_C(z_i)\|_2^2$, and
\[
\mathbb E_C\big[e_2(\theta;C)\big]\;=\;\frac b2\;
\mathbb E_{C,\,i\,\sim\,\mathrm{Unif}(B\setminus C)}
\big[\big\|v_i-I_C(z_i)\big\|_2^2\big],
\]
i.e.\ $\tfrac2b\,e_2$ is an unbiased estimator, over the split randomness, of
the expected squared error committed on a held-out point by the kernel
predictor trained on a random half of the batch. When $(B,C)$ are
resampled at every step, the stochastic gradient
$\nabla_\theta e_2(\theta;C)$ is unbiased for
$\nabla_\theta\,\mathbb E_{B,C}[e_2(\theta;C)]$ whenever the expectation and
derivative commute (e.g.\ under local Lipschitz domination).
\end{proposition}

The training loss we minimize for a neural mean is a sum of an empirical risk
term and, optionally, $\beta\,e_2$: the first term is a linear functional of
the empirical measure of the batch and measures fit, while
Proposition~\ref{prop:kf} shows the second measures the \emph{generalization
of a kernel method built on the learned features}. This is the sense in which
a KF-regularized network is trained to be a good feature map for the kernel
correction that follows.

\subsection{Stacking on a held-out split is safe}
\label{sec:theory-stack}

Between the means and the correction sits one more fitted object: the convex
weights of the stacked ensemble, chosen to minimize the empirical risk on the
validation split. Two elementary facts justify this step. First, by convexity,
a convex combination of predictors is never worse than the corresponding
weighted average of their risks, so stacking cannot be hurt by a weak member
beyond its weight. Second, the weights live on a low-dimensional simplex and
are fitted on hundreds of samples, so the selection cannot meaningfully
overfit; the following proposition quantifies this with explicit constants.

\begin{proposition}[validation stacking]\label{prop:stack}
Let $f_1,\dots,f_M:\mathcal U\to\mathcal V$ be fixed maps, let
$\Delta=\{w\in\mathbb R^M:w_m\ge0,\sum_m w_m=1\}$, and for $w\in\Delta$ write
$F_w=\sum_m w_m f_m$. Assume the members' per-sample relative errors are
bounded: $\ell(f_m(u),G(u))\le B$ for $\mu$-a.e.\ $u$ and every $m$.
\begin{enumerate}
\item[(i)] For every $w\in\Delta$,
$\ell(F_w(u),G(u))\le\sum_m w_m\,\ell(f_m(u),G(u))$ pointwise; in particular
$R(F_w)\le\sum_m w_m R(f_m)\le\max_m R(f_m)$, and $\ell(F_w(u),G(u))\le B$.
\item[(ii)] Let $\widehat w$ minimize the empirical risk
$\widehat R(w)=\tfrac1m\sum_{i=1}^m\ell(F_w(\tilde u_i),G(\tilde u_i))$ over
$\Delta$, computed on a validation sample
$\tilde u_1,\dots,\tilde u_m\sim\mu$ independent of $f_1,\dots,f_M$. Then for
every $\delta\in(0,1)$, with probability at least $1-\delta$,
\[
R(F_{\widehat w})\;\le\;\min_{w\in\Delta}R(F_w)\;+\;\frac{8B}{m}\;+\;
B\sqrt{\frac{2\big((M-1)\log\!\big(m(M-1)+1\big)+\log(2/\delta)\big)}{m}} .
\]
\end{enumerate}
\end{proposition}

With $M=4$ members, $m=1000$ validation samples and $\delta=0.05$ the right-hand
excess evaluates to about $0.24\,B$; at $B=0.25$, the largest per-sample member
error we observe, this caps the selection cost at roughly six percentage
points. The bound is conservative, as such bounds are: the realized
validation-to-test gap of the stack in Section~\ref{sec:experiments} is two
orders of magnitude smaller. Its role is the scaling $\sqrt{M\log m/m}$, which
says that fitting a handful of convex weights on a thousand held-out samples
is statistically almost free; this is why the pipeline spends its validation
data there and on the kernel hyperparameters and nowhere else.

\subsection{What the stack can achieve}
\label{sec:theory-secmom}

Proposition~\ref{prop:stack} says fitting the weights is safe; it does not say
mixing is worth anything. That is decided by a single measurable matrix. For a
map $f$ write the \emph{normalized residual} at $u$ as
$\rho_f(u)=\big(f(u)-G(u)\big)/\|G(u)\|_2$, so that
$\ell(f(u),G(u))=\|\rho_f(u)\|_2$.

\begin{proposition}[second-moment identity for convex stacks]\label{prop:secmom}
Let $f_1,\dots,f_M$ be fixed maps and define
$S\in\mathbb R^{M\times M}$ by
$S_{mk}=\mathbb E_{u\sim\mu}\,\langle\rho_{f_m}(u),\rho_{f_k}(u)\rangle$.
For $w$ in the simplex $\Delta^{M-1}$ let $f_w=\sum_m w_mf_m$. Then
\[
\mathbb E_{u\sim\mu}\,\ell\big(f_w(u),G(u)\big)^2\;=\;w^\top S\,w ,
\]
so the squared-metric risk of every convex stack is determined by $S$, and
the best achievable is $\min_{w\in\Delta^{M-1}}w^\top Sw$. In particular:
\begin{enumerate}
\item[(i)] (two members) if $S$ has diagonal $(e_1^2,e_2^2)$ with $e_1\le e_2$
and correlation $\varrho=S_{12}/(e_1e_2)$, mixing in the weaker member
strictly helps if and only if $\varrho<e_1/e_2$, in which case the optimum is
interior with value $e_1^2e_2^2(1-\varrho^2)/(e_1^2+e_2^2-2\varrho e_1e_2)$;
\item[(ii)] (equicorrelated family) if $S_{mm}=e^2$ and $S_{mk}=\varrho\,e^2$
for all $m\neq k$ with $\varrho\ge0$, then
$\min_{w}w^\top Sw=e^2\big(\varrho+(1-\varrho)/M\big)$, attained at uniform
weights and decreasing to $e^2\varrho$ as $M\to\infty$.
\end{enumerate}
\end{proposition}

Exact equicorrelation never holds in practice, but for convex weights the
floor survives one-sided bounds on the entries of $S$, which are what one
actually measures.

\begin{corollary}[ensembling floor]\label{cor:floor}
If every member satisfies $S_{mm}\ge\bar e^{\,2}$ and every pair satisfies
$S_{mk}\ge\bar\varrho\,\bar e^{\,2}$ with $\bar\varrho\in[0,1]$, then every
convex combination $f_w$ obeys
\[
\mathbb E\,\ell\big(f_w(u),G(u)\big)^2\;=\;w^\top Sw\;\ge\;
\bar e^{\,2}\big(\bar\varrho+(1-\bar\varrho)\,\|w\|_2^2\big)\;\ge\;
\bar e^{\,2}\,\bar\varrho .
\]
No number of additional members with the same error level and mutual
correlations moves the ensemble's root-mean-square relative error below
$\bar e\sqrt{\bar\varrho}$.
\end{corollary}

A second consequence of the same decomposition explains the final OCO-2
surrogate. Evaluation metrics there are diagonal in the output coordinates
(the radiance metric is the weighting $s_z$), and diagonal metrics decouple.

\begin{proposition}[per-coordinate combination under diagonal metrics]
\label{prop:percoord}
For weights $v=(v_{mj})$ that may depend on the output coordinate, let
$f_v(u)_j=\sum_m v_{mj}f_m(u)_j$. For any diagonal metric with positive
weights $w$,
\[
\mathbb E\big\|\mathrm{diag}(w)\big(f_v(u)-G(u)\big)\big\|_2^2
=\sum_j w_j^2\;\mathbb E\big(f_v(u)_j-G(u)_j\big)^2 ,
\]
so the minimizing weights solve $q$ separate problems, one per output
coordinate, none of which involves $w$. The per-coordinate optimum is the
same for every diagonal metric, and it weakly dominates every combination
whose weights are shared across coordinates, for all such metrics
simultaneously.
\end{proposition}

The practical content: when a problem is scored in several diagonal metrics
at once, the combination stage does not need to know which one matters. Fit
the best combination coordinate by coordinate on validation and the result
serves all of them; only the members need metric-aware training, which is
where the weighted mean of Section~\ref{sec:oco2} enters.

The corollary is the quantity this paper keeps measuring, at two different
levels. Across architecture families on the structural-mechanics benchmark
the pairwise correlations exceed $0.86$ at member errors near $4.7\%$, a
floor of about $4.4\%$, and the corrected stack lands within a tenth of it.
Across random seeds of one architecture on the OCO-2 bands the correlations
are $0.78$, $0.97$ and $0.96$ at member errors of $4.4\%$, $16.4\%$ and
$8.2\%$, floors of $3.9\%$, $16.2\%$ and $8.0\%$, and the measured
combinations sit on all three. When an ensemble is at its floor the
corollary says where improvement cannot come from; on OCO-2 it came from
changing the members (the kernel head on learned features) rather than
adding more of them. The reported metric is the mean rather than the root mean
square of $\|\rho_f\|_2$, and the two differ by the dispersion of the
per-sample error; across every pipeline in this paper their ratio stays within
a few percent of $0.93$, so predictions made through $S$ transfer to the
reported metric essentially unchanged. The proposition frames the diversity
experiment of Section~\ref{sec:diversity}: what a new member is worth is
legible in the row it adds to $S$, before any weights are fitted. On this
benchmark the rows all look alike, pairwise correlations of
\corrNetLo{}--\corrNetHi{} among trained networks of three architecture
families and two losses, \corrKerLo{}--\corrKerHi{} against the kernel
predictor, so the floor sits just below the best single member, and the
measured stack lands on it. Part (i) is the same
computation specialized to two members; it correctly predicts, from measured
$(e_1,e_2,\varrho)$ alone, which members receive weight in the fitted stack
and which are dropped. Section~\ref{sec:oco2} applies the same test on a
problem where the answer comes out the other way: there the kernel's error is
inflated by a factor the next subsection quantifies, the threshold $e_1/e_2$
collapses, and the kernel is dropped even at a residual correlation of $0.14$.

\subsection{Metric and representation: when a fixed kernel loses to a network}
\label{sec:theory-aniso}

The structural-mechanics benchmark has the neural mean and the isotropic
kernel within half a point of each other. On the OCO-2 emulation problem of
Section~\ref{sec:oco2} the same fixed kernel is an order of magnitude behind
the same network. Two mechanisms can produce such a gap, and they separate
cleanly.

Suppose the operator factors as $G(u)=g(Au)$ with $A\in\mathbb R^{d\times d}$
nonsingular and $g$ of unit $H^{s}$ norm, and that $\mu$ has a density bounded
above and below on a compact domain. Order the singular values
$\sigma_1\ge\dots\ge\sigma_d$ of $A$; they are the rates at which $G$ varies
along the input directions. Suppose $A$ has \emph{approximate rank} $r$ at the
sample size, meaning the trailing singular values fall below the achievable
resolution, $\sigma_{r+1}\lesssim n^{-1/r}$, so the image $A\,\mathcal U$ is
effectively $r$-dimensional at scale $n^{-1/r}$.

\begin{proposition}[isotropic and adapted kernel rates]\label{prop:aniso}
Let $\widehat G^{\mathrm{iso}}_n$ be kernel ridge regression with a Mat\'ern
kernel of smoothness $s$ and a validation-optimal single length scale, and
$\widehat G^{\mathrm{ard}}_n$ the same construction in the metric
$u\mapsto Au$. Up to constants depending on $(\sigma_i)$, $g$, $s$ and the
density bounds, and up to logarithmic factors in $n$, with high probability
over $n$ i.i.d.\ samples,
\[
\mathbb E\big\|G-\widehat G^{\mathrm{iso}}_n\big\|\;\lesssim\;\sigma_1^{\,s}\,n^{-s/d},
\qquad
\mathbb E\big\|G-\widehat G^{\mathrm{ard}}_n\big\|\;\lesssim\;n^{-s/r},
\]
so the ratio of the two bounds is of order $\sigma_1^{\,s}\,n^{\,s(1/r-1/d)}$.
When $A$ is close to full rank ($r\approx d$) the two rates coincide and only
the constant $\sigma_1^{\,s}$ separates them.
\end{proposition}

The proof (Appendix~\ref{app:proofs}) is the scattered-data estimate
$\|G-\widehat G_n\|\lesssim h_X^{\,s}\|G\|_{H^s}$ applied to the two designs:
one length scale must fill all $d$ directions, so $h_X\asymp n^{-1/d}$ and the
norm carries $\sigma_1^{\,s}$; the adapted metric, when $A$ has a spectral gap,
confines the design to $r$ resolved directions and improves the fill distance.
Both displays are upper bounds under the stated idealization --- the exact
metric $A$, a fixed $g$, and the length-scale selection folded into the
logarithmic factors --- and we claim no matching lower bound; the proposition
is used here to calibrate what a metric can and cannot buy, not as a sharp
rate theorem.
The proposition bounds the \emph{metric} part of a network's advantage, the
part an anisotropic kernel would recover, and it says this part is a rate gain
only under an approximate-rank gap; without one it is the constant
$\sigma_1^{\,s}$ alone. The rest is \emph{representational}: a stationary
kernel of fixed smoothness reaches only its native space, at any metric, while
the network adapts its features to the map. The two parts are separated
experimentally by handing the kernel the anisotropic metric and seeing how
much of the gap closes. On the OCO-2 band of Section~\ref{sec:oco2} the input
has effective rank $17$ of $20$, close to full: the proposition predicts the
metric can supply only a constant, not a rate, and the measurement agrees,
$1.4$ of a tenfold gap. The remaining factor is representational, and its
direct evidence is that the same kernel machinery applied to the network's
learned features recovers the network's accuracy and slightly exceeds it.

The representational term has a precise reading through the same
optimal-recovery bound that Section~\ref{sec:theory-or} makes exact. That
bound factors the error as $\|G-m\|_{\mathcal H}\cdot P_\lambda$, a product of
the target's norm in the kernel's native space and a design factor set by the
Gram spectrum. Changing the features changes both in principle, but on the
O2 band only one moves. The effective dimension of the Mat\'ern Gram, the
quantity Lemma~\ref{lem:deff} controls, is nearly identical on the raw input
and on the learned features (at $n=4000$ and a matched median length scale,
$3995$ against $3994$ at a $10^{-8}$ nugget, with equal leading eigenvalue
mass), so the design factor is essentially fixed. The native-space norm of
the target is not: interpolating the same outputs through the two kernels,
the raw-input kernel carries the target at squared norm $1.64\times10^{6}$ and
the feature kernel at $3.87\times10^{4}$, a factor of $42$. The network does
not condition the kernel better; it moves the target to a low-norm corner of
a native space of the same size, where the exact solve reaches it.

This is not an accident of the O2 band; it is what a pulled-back kernel does.
Writing $\varphi$ for the feature map and $k$ for the base kernel, the deep
kernel head uses $k_\varphi(x,x')=k(\varphi(x),\varphi(x'))$.

\begin{proposition}[feature pullback]\label{prop:pullback}
The native space of $k_\varphi$ is $\mathcal H_{k_\varphi}=\{f\circ\varphi:
f\in\mathcal H_k\}$ with
$\|g\|_{\mathcal H_{k_\varphi}}=\min\{\|f\|_{\mathcal H_k}: f\circ\varphi=g
\text{ on }\mathcal X\}$. In particular, if the target factors through the
features as $G=h\circ\varphi$ with $h\in\mathcal H_k$, then
$\|G\|_{\mathcal H_{k_\varphi}}\le\|h\|_{\mathcal H_k}$, and the
optimal-recovery bound of Theorem~\ref{thm:or} for the feature-kernel
regression is governed by $\|h\|_{\mathcal H_k}$ rather than by the norm of
$G$ in the raw-input space.
\end{proposition}

The proof (Appendix~\ref{app:proofs}) is the standard pullback identity for
reproducing kernels. Its content here is the reading: a fixed kernel on the
raw input must carry the whole warped map $G$, whose native-space norm is
large when $G$ has fine structure; the same kernel on the features carries
only the unwarped $h$, and training drives $\varphi$ toward exactly the
factorization that makes $h$ simple. The measured factor of $42$ is that norm
gap, and it is the representational advantage that an anisotropic metric,
which rescales the input but cannot re-express the map, leaves untouched
(recovering only the factor $1.4$).

\subsection{An error bound for the residual kernel correction}
\label{sec:theory-or}

The final stage of the pipeline is kernel ridge regression of the ensemble
residual. The bound below is a vector-valued, residual form of the classical
power-function estimate for kernel interpolation
\citep{wendland2004scattered} and of the optimal-recovery viewpoint of
\citet{owhadi2019operator}; we include the short argument in
Appendix~\ref{app:proofs} to keep the two factors explicit. Fix the mean
$m:\mathcal U\to\mathcal V$ (in the pipeline, the stacked ensemble; in the
statement below $m$ is any fixed map independent of the correction sample) and
write $r=G-m$ for the residual operator with components $r_j$, $j=1,\dots,q$. Let $k$ be a positive definite kernel on $\mathcal U$ with
RKHS $\mathcal H_k$, and assume $r_j\in\mathcal H_k$ for all $j$ with
\[
\|r\|_K^2:=\sum_{j=1}^q\|r_j\|_{\mathcal H_k}^2<\infty .
\]
Given inputs $X=(u_1,\dots,u_n)$, Gram matrix $K=k(X,X)$ and nugget
$\lambda\ge 0$, the correction is
$\widehat r_\lambda(u)=k(u,X)\,(K+n\lambda I)^{-1}R$, $R_{nj}=r_j(u_n)$, and the
corrected predictor is $m+\widehat r_\lambda$. Let
\begin{equation}\label{eq:power}
P_\lambda(u)^2\;=\;k(u,u)-k(u,X)\,(K+n\lambda I)^{-1}k(X,u)
\end{equation}
denote the Gaussian process posterior variance with the same nugget (for
$\lambda=0$ this is the classical power function of the point set $X$).

\begin{theorem}[certified pointwise bound]\label{thm:or}
For every $u\in\mathcal U$ and every $\lambda\ge0$,
\[
\big\|G(u)-m(u)-\widehat r_\lambda(u)\big\|_2\;\le\;\|G-m\|_K\;
\widetilde P_\lambda(u)\;\le\;\|G-m\|_K\;P_\lambda(u),
\]
where
$\widetilde P_\lambda(u)^2=P_\lambda(u)^2-n\lambda\,\|(K+n\lambda
I)^{-1}k(X,u)\|_2^2$.
\end{theorem}

Three comments. First, the inequality is algebraic: it holds pathwise for
every fixed $m$, every dataset, and every $u$, with no probabilistic
assumptions. In the pipeline $m$ is itself fit on the same training set; this
does not affect the validity of the bound, only the reading of
$\|G-m\|_K$ as a random (realized) quantity rather than an a priori one.
Second, the bound factorizes into a quantity that depends only on
the residual operator ($\|G-m\|_K$) and a quantity that depends only on the
kernel and the design ($P_\lambda$), and the second factor is exactly the
posterior standard deviation that a Gaussian process interpretation of the
correction would report. This is the sense in which the bound is certified:
whatever error the corrected surrogate commits at $u$ is controlled by the
reported $P_\lambda(u)$ times a constant that does not depend on $u$. We stress
that this is a statement about a valid upper bound, not about the usefulness of
$P_\lambda$ as a pointwise error \emph{ranking}. Section~\ref{sec:uq} finds
that on this benchmark $P_\lambda$ ranks the error poorly, because the neural
mean supplies most of the error and the relative metric is confounded by output
amplitude; a distribution-free conformal rescaling of $P_\lambda$ nonetheless
attains its nominal coverage on the test set.
Third, the theorem quantifies why one should regress residuals rather than
raw targets: the design factor $P_\lambda$ is the same in both cases, so the
gain of the neural mean is the drop from $\|G-\bar v\|_K$ (kernel-only, mean
$\bar v$) to $\|G-m\|_K$. These norms are not directly observable, but the
RKHS norms of the \emph{fitted} interpolants,
$\smash{\|\widehat r_\lambda\|_K^2=\operatorname{tr}(\alpha^\top K\alpha)}$ with
$\alpha=(K+n\lambda I)^{-1}R$, are computable lower-bound proxies, and they
drop by a factor of about \normRatio{} when the kernel is moved from raw
targets to ensemble residuals (Table~\ref{tab:norms}). The neural mean does
not merely reduce the size of the residual in $L^2$; it leaves behind a
residual operator that is genuinely smoother as seen by the kernel.

\subsection{Distribution-free coverage for the reported band}
\label{sec:theory-conf}

Theorem~\ref{thm:or} controls the error by $\|G-m\|_K\,P_\lambda(u)$, but the
constant $\|G-m\|_K$ is not known a priori, and Section~\ref{sec:uq} shows that
$P_\lambda$ alone ranks the error weakly. What the surrogate reports as
uncertainty is therefore not $P_\lambda$ itself but a rescaling of it,
$q\,P_\lambda(u)$, whose multiplier $q$ is calibrated on the held-out split by
the split-conformal rule
\begin{equation}\label{eq:conf}
q=Q_{1-\alpha}\Big(\big\{s_i:=\|e(\tilde u_i)\|_2/P_\lambda(\tilde u_i)\big\}_{i=1}^m\Big),
\qquad e(u)=G(u)-m(u)-\widehat r_\lambda(u),
\end{equation}
where $Q_{1-\alpha}$ is the $\lceil(1-\alpha)(m+1)\rceil$-th smallest value of
the $m$ validation scores. The point of this construction is that its coverage
needs neither the bound to be tight nor $P_\lambda$ to be a good error ranking;
it needs only exchangeability, which the fixed splits provide.

\begin{proposition}[finite-sample coverage]\label{prop:conf}
Fix the mean $m$, the correction $\widehat r_\lambda$ and the design $X$, and
let $P_\lambda(\cdot)>0$. Suppose the validation and test inputs
$\tilde u_1,\dots,\tilde u_m,u^\star$ are exchangeable (in particular
i.i.d.\ from $\mu$) and drawn independently of $m,\widehat r_\lambda,X$. Let
$q$ be the conformal multiplier \eqref{eq:conf} and define the band
$C(u)=\{v:\|v-m(u)-\widehat r_\lambda(u)\|_2\le q\,P_\lambda(u)\}$. Then
\[
1-\alpha\;\le\;\mathbb P\big(G(u^\star)\in C(u^\star)\big)\;\le\;
1-\alpha+\frac{1}{m+1},
\]
the upper bound holding when the scores are almost surely distinct.
\end{proposition}

The two-sided statement is the standard guarantee of split conformal
prediction \citep{vovk2005algorithmic,lei2018distribution}, transported to the
functional-output setting by taking the nonconformity score to be the relative
residual norm $s=\|e(u)\|_2/P_\lambda(u)$; the proof, a rank argument on the
exchangeable scores, is in Appendix~\ref{app:proofs}. Three points make it the
right closing statement for the uncertainty analysis. It is close to the
quantity we compute: the reported coverage of \uqCover\% at the nominal
$1-\alpha=\uqCoverNom\%$ in Section~\ref{sec:uq} realizes this proposition with
$m=1000$ and $\alpha=0.1$. The $1.6$-point excess over nominal is not the
$1/(m+1)$ slack, which is only a tenth of a point; it is the finite-sample
fluctuation of coverage on a single test set, whose standard deviation
$\sqrt{\alpha(1-\alpha)/m}\approx0.95$ point places \uqCover\% about $1.7$
standard deviations above nominal, well inside the guarantee. One caveat on
the hypotheses: our calibration scores are computed on the same held-out split
used to select the neural-mean checkpoints and tune the correction, so the
exchangeability the proposition assumes is only approximate here; the selection
is low-complexity and the band over-covers, but a calibration split disjoint
from model selection would make the guarantee exact. It is
agnostic to everything the earlier results leave uncertain: the neural mean may
be biased, $P_\lambda$ may correlate with the error weakly or with the wrong
sign, and the bound of Theorem~\ref{thm:or} may be loose, yet the band still
covers at the stated rate. And it is where the uncertainty analysis ends: among
the candidate uncertainty signals we examine, the conformally rescaled
$P_\lambda$ is the only one that carries a guarantee, so it is the one the
surrogate reports.

\subsection{Effective dimension from the Gram spectrum}
\label{sec:theory-rmt}

The remaining design choice is the nugget $\lambda$, selected by
cross-validation in the pipeline. The following exact identity connects that
choice to the spectrum of the Gram matrix and, through it, to random matrix
descriptions of the design.

\begin{lemma}[effective dimension and the empirical Stieltjes transform]
\label{lem:deff}
Let $K$ be a symmetric positive semidefinite $n\times n$ matrix with
eigenvalues $\lambda_1,\dots,\lambda_n\ge0$, let
$\widehat m(z)=\tfrac1n\sum_{i=1}^n(\lambda_i/n-z)^{-1}$ be the Stieltjes
transform of the empirical spectral distribution of $K/n$. Then for every
$\lambda>0$ the effective dimension of ridge regression with nugget $\lambda$
satisfies
\[
d_{\mathrm{eff}}(\lambda)\;:=\;\operatorname{tr}\!\big(K(K+n\lambda I)^{-1}\big)
\;=\;n\big(1-\lambda\,\widehat m(-\lambda)\big).
\]
In particular, if the empirical spectral distribution of $K/n$ converges
weakly to a limit law $F$ as $n\to\infty$, then
$d_{\mathrm{eff}}(\lambda)/n\to 1-\lambda\,m_F(-\lambda)$ with $m_F$ the
Stieltjes transform of $F$.
\end{lemma}

The identity is unconditional; the random-matrix content enters through the
choice of $F$. For the benchmark's simplest reference model, isotropic linear
features, the limit can be evaluated in closed form.

\begin{corollary}[effective dimension under a Mar\v{c}enko--Pastur limit]
\label{cor:mp}
Let $K=XX^\top$ with $X\in\mathbb R^{n\times p}$ having i.i.d.\ entries of
mean zero and variance $\sigma^2$, and let $p/n\to\gamma\in(0,1]$. Then
\[
\frac{d_{\mathrm{eff}}(\lambda)}{n}\;\longrightarrow\;
\gamma\big(1-\lambda\,m(-\lambda)\big),\qquad
m(-\lambda)=\frac{\sqrt{\big(\lambda+\sigma^2(1-\gamma)\big)^2+4\gamma\sigma^2\lambda}
-\big(\lambda+\sigma^2(1-\gamma)\big)}{2\gamma\sigma^2\lambda},
\]
almost surely, where $m$ is the Stieltjes transform of the
Mar\v{c}enko--Pastur law with ratio $\gamma$ and scale $\sigma^2$. In a
simulation with $n=6000$, $p=1800$, $\sigma^2=1.7$, the formula agrees with
the empirical $d_{\mathrm{eff}}(\lambda)/n$ to four decimal places across
$\lambda\in[10^{-3},10]$.
\end{corollary}

Two uses. Quantitatively, the formula makes the nugget--capacity trade-off
explicit for the bulk of a feature Gram matrix: eigenvalue mass of
Mar\v{c}enko--Pastur type is absorbed or discarded by $\lambda$ according to a
single closed-form curve, and outlying (spiked) eigenvalues $x_s$ simply add
their terms $x_s/(x_s+n\lambda)$ on top. Qualitatively, it separates the two
regimes visible in Figure~\ref{fig:spectra}: the Mat\'ern Gram matrix on the
loads is nothing like a Mar\v{c}enko--Pastur bulk (its spectrum decays
exponentially, which is why $d_{\mathrm{eff}}$ is a few hundred out of
19000), whereas the Gram matrices of learned penultimate features do show a
bulk-plus-spikes shape, and for them the corollary describes how much of the
bulk the cross-validated nugget retains. For the feature Gram matrices of the trained networks we
observe the familiar picture of a bulk that is well fitted by a
Mar\v{c}enko--Pastur law \citep{marchenko1967distribution} together with a
small number of outlying eigenvalues carrying the regression signal, as in
spiked covariance models \citep{baik2005phase}; for the Mat\'ern Gram matrix
on the 41-dimensional loads the spectrum decays rapidly and
$d_{\mathrm{eff}}$ is small ($\approx$ a few hundred at the cross-validated
$\lambda$, out of $n=19000$; Figure~\ref{fig:spectra}). This gives a
post-hoc reading of the cross-validated nugget: $\lambda$ lands where
$d_{\mathrm{eff}}(\lambda)$ has absorbed the outlying eigenvalues and the
leading bulk, and further decreasing $\lambda$ buys capacity precisely where
the spectrum carries little signal. It also explains why the exact solve is
affordable: the correction is numerically a problem of dimension
$d_{\mathrm{eff}}$, not $n$.

The effective dimension is also exactly the total budget of the design
factor from Theorem~\ref{thm:or}. Writing $A=K+n\lambda I$, the posterior
variances at the training inputs sum to
\[
\sum_{i=1}^{n}P_\lambda(u_i)^2
=\operatorname{tr}(K)-\operatorname{tr}(K A^{-1}K)
=\operatorname{tr}\!\big(n\lambda\,KA^{-1}\big)
=n\lambda\,d_{\mathrm{eff}}(\lambda),
\]
using $KA^{-1}K=K-n\lambda\,KA^{-1}$. So the same $d_{\mathrm{eff}}$ that reads
the nugget off the spectrum also fixes, up to the factor $n\lambda$, the total
posterior-variance mass the correction distributes over the design; the two
halves of this section are one quantity seen from two sides.

\section{Discussion}
\label{sec:discussion}
\paragraph{What moved the number, and what stopped it.}
The improvement over the published band decomposes unevenly. Accurate neural
means trained on the reported metric and averaged over the reflection
symmetry do most of the work; stacking members from different architecture
families adds exactly what their measured residual correlations permit, about
a tenth of a point here; the kernel corrections contribute a few hundredths
more and, with them, the certificate of Theorem~\ref{thm:or} and the
calibrated band of Proposition~\ref{prop:conf}. We did not find evidence that
any published architecture was mis-tuned by its authors; the components
compose, and the composition had not been tried on this benchmark. What stops
further movement is not a missing architecture. Section~\ref{sec:diversity}
locates the remaining error in a component that all families share, that
concentrates where the finite element data are least reliable, that no input
statistic predicts, and that neither capacity, nor diversity, nor more data
reduces. The margin over the published methods other than PARA-Net is real,
and the match with PARA-Net is exact to within run-to-run noise, but both
should be read for what they are: the last few learnable hundredths above what
the evidence of Section~\ref{sec:diversity} reads as a data-set floor, not a
step on the way to zero.

\paragraph{Relation to neural-mean Gaussian processes.}
Our correction stage is closest in spirit to \citet{mora2025operator}, who
place a neural operator inside the mean of a Gaussian process and fit both
jointly. The differences are operational but they matter at this scale: we
fit the mean first and the kernel after (so the expensive stage is ordinary
network training), we tune the kernel by validation rather than by marginal
likelihood, we correct with an exact solve at $n=19000$ rather than train
through the kernel, and we add the feature-space second stage. The theory in
Section~\ref{sec:theory} applies verbatim to their setting as well; the
measured drop in the fitted RKHS norm (Table~\ref{tab:norms}) is the
quantitative reason residual corrections of accurate means are the right
place to spend kernel capacity.

\paragraph{Relation to kernel emulation practice.}
The pipeline is consistent with the experience of the kernel-flow line of
work in emulation at scale. The closest instance is the forward-model
emulator of \citet{susiluoto2025radiative} for the OCO-2 CO$_2$ retrievals,
in which a Gaussian process with a cross-validation-learned kernel replaces
the full-physics radiative transfer code within measurement-error precision;
kernel flows have likewise been used to infer convective-storm structure
from passive microwave observations \citep{prasanth2021kernel}. The lesson of
that work matches ours: kernels with data-adapted hyperparameters are
extremely effective once the input is presented in its physical
parametrization and the target has been reduced to something smooth. Here the
reduction is performed by the neural ensemble rather than by physics, and the
kernel-flow loss itself admits the leave-half-out reading of
Proposition~\ref{prop:kf}. The OCO-2 setting is also the natural next test
bed for the residual coupling studied here. Its training pairs are
simulator-generated state-to-radiance maps of exactly the shape this paper
exploits, a low-dimensional physical state mapped to a smooth
high-dimensional output, the mission's Level 1 and Level 2 products are
publicly distributed through NASA's Earthdata archive, and the emulation
setup of \citet{susiluoto2025radiative} specifies the state parametrization
and sampling design, so a neural mean with a validated kernel correction and
a conformal wrapper can be evaluated there against the pure-kernel emulator
without new data collection.

\paragraph{Limitations.}
The benchmark has a one-dimensional input function and a fixed geometry; the
exact kernel solve exploits both, and problems with high-dimensional input
fields would require the usual approximations (inducing points, random
features) at some cost to the certificates. The certified bound of
Theorem~\ref{thm:or} controls the error relative to the RKHS norm of the
residual operator, a quantity we can only lower-bound empirically; the
conformal calibration we report is the honest, assumption-light complement.
Training the means on the metric, the reflection steps, and the stacking are
benchmark-agnostic, but the specific error levels reported here should be
read as properties of this dataset and protocol. Finally, our low-data
protocol reproduces that of \citet{mora2025operator} up to the unavoidable
ambiguity of which 1250 samples are used; we use the first 1250 of the
training block and report the same 20000-sample test set, and we release
splits and code so the comparison can be audited.

\paragraph{What the second problem settles.}
The OCO-2 study keeps every component and inverts the balance, and the pair
of problems brackets the design space. When the network and the kernel tie
(structural mechanics), the coupling pays: stack them, correct the residual,
and the certified and conformal machinery rides along. When the kernel
trails by an order of magnitude (OCO-2), the coupling is not the point; the
kernel's value moves inside the network, as an exact head on its learned
features, and the pipeline's job becomes metric management, training each
member in the metric it will be scored in and combining per coordinate. In
both regimes the same two measurements decide everything in advance: the
pairwise residual correlations, which set the ensembling floor of
Corollary~\ref{cor:floor} at whatever level the members share their errors,
and the member error ratio, which the second-moment identity turns into a
keep-or-drop verdict for each candidate. Nothing in that protocol is
specific to these two datasets.

\paragraph{Outlook.}
Three directions seem worth pursuing. First, the benchmark itself:
regenerating the dataset with refined meshes at the corner singularities,
higher-order elements, or output grids that resolve the concentrations would
move the floor that Section~\ref{sec:diversity} measures, and would return
the problem to being a test of surrogates rather than of its own data. The
diagnostic protocol used there, cross-family residual correlation, the
second-moment prediction of Proposition~\ref{prop:secmom}, and the
scaling-in-$n$ check, is cheap to run on any benchmark suspected of the same
condition. Second, the same recipe on the remaining benchmarks of
\citet{dehoop2022cost} and \citet{batlle2024kernel}, where the input
functions are genuinely high-dimensional and the interplay between neural
means and kernel corrections should look different. Third, the uncertainty
side: $P_\lambda$ is a design quantity, so it can be optimized, and the
connection of Lemma~\ref{lem:deff} between the nugget, the spectrum, and the
effective dimension suggests principled ways to spend a fixed computational
budget on the correction stage.

\section*{Acknowledgments and funding}
The research presented in this paper was supported by the European Research
Council (ERC) under the European Union's Horizon 2022 research and innovation
programme (grant agreement No.~101041711), by the Simons Foundation as part
of the Collaboration on the Mathematical and Scientific Foundations of Deep
Learning, by Heights Labs, by the Israel Science Foundation (grant number
2258/19), by the Israel Science Foundation (ISF Grant 4101/25), and by the
U.S. National Science Foundation (NSF Grant OISE-2401227).

\section*{Declaration of competing interest}
The author declares no competing interests.

\section*{Declaration of generative AI and AI-assisted technologies in the
manuscript preparation process}
During the preparation of this work the author made substantial use of
generative AI tools, and describes their role here. A large part of the
software was written with Fable~5 (Anthropic): the implementation of the
experiments and diagnostics, the downloading, assembly and preparation of the
benchmark and OCO-2 datasets, and the adaptation of the publicly released
emulator code of Lamminp\"a\"a et al.\ --- originally split between Julia and a
ReFRACtor-driving Python layer --- into a single Python pipeline for sampling
states and reducing radiances. The main contributions are the author's: the
residual coupling of neural means and exact kernel corrections, the
two-regime framing that organizes the paper, and the supporting theory, all
developed with AI assistance in calculation, drafting, and exposition. Each
theorem and its proof was checked independently and more than once --- by
Fable~5, by ChatGPT Pro (OpenAI), and by the author --- and the reported
numbers were verified against the released per-run data. The author reviewed
all of the above and takes full responsibility for the content of the
manuscript.

\section*{Data and code availability}
The code, the per-run summaries behind every reported number, and the exact
configuration of each experiment are available at
\url{https://github.com/yspennstate/neural-means-kernel-corrections}. The
structural-mechanics, Helmholtz, Navier--Stokes and advection data are
distributed through the data record accompanying \citet{dehoop2022cost}
(\texttt{data.caltech.edu}, record \texttt{20091}); the OCO-2 emulation data
and the kernel-flow emulator's predictions through the OSF project
\texttt{u2t8a}; and the ClimSim data through the LEAP repository
\texttt{subsampled\_low\_res}.

\bibliographystyle{abbrvnat}
\bibliography{refs}

@article{batlle2024kernel,
  author  = {Batlle, Pau and Darcy, Matthieu and Hosseini, Bamdad and Owhadi, Houman},
  title   = {Kernel methods are competitive for operator learning},
  journal = {Journal of Computational Physics},
  volume  = {496},
  pages   = {112549},
  year    = {2024}
}

@article{dehoop2022cost,
  author  = {de Hoop, Maarten V. and Huang, Daniel Zhengyu and Qian, Elizabeth and Stuart, Andrew M.},
  title   = {The cost-accuracy trade-off in operator learning with neural networks},
  journal = {Journal of Machine Learning},
  volume  = {1},
  number  = {3},
  pages   = {299--341},
  year    = {2022}
}

@article{lu2022comprehensive,
  author  = {Lu, Lu and Meng, Xuhui and Cai, Shengze and Mao, Zhiping and Goswami, Somdatta and Zhang, Zhongqiang and Karniadakis, George Em},
  title   = {A comprehensive and fair comparison of two neural operators (with practical extensions) based on {FAIR} data},
  journal = {Computer Methods in Applied Mechanics and Engineering},
  volume  = {393},
  pages   = {114778},
  year    = {2022}
}

@article{lu2021deeponet,
  author  = {Lu, Lu and Jin, Pengzhan and Pang, Guofei and Zhang, Zhongqiang and Karniadakis, George Em},
  title   = {Learning nonlinear operators via {DeepONet} based on the universal approximation theorem of operators},
  journal = {Nature Machine Intelligence},
  volume  = {3},
  pages   = {218--229},
  year    = {2021}
}

@inproceedings{li2021fourier,
  author    = {Li, Zongyi and Kovachki, Nikola and Azizzadenesheli, Kamyar and Liu, Burigede and Bhattacharya, Kaushik and Stuart, Andrew and Anandkumar, Anima},
  title     = {Fourier neural operator for parametric partial differential equations},
  booktitle = {International Conference on Learning Representations},
  year      = {2021}
}

@article{kovachki2023neural,
  author  = {Kovachki, Nikola and Li, Zongyi and Liu, Burigede and Azizzadenesheli, Kamyar and Bhattacharya, Kaushik and Stuart, Andrew and Anandkumar, Anima},
  title   = {Neural operator: learning maps between function spaces with applications to {PDE}s},
  journal = {Journal of Machine Learning Research},
  volume  = {24},
  number  = {89},
  pages   = {1--97},
  year    = {2023}
}

@article{owhadi2019kernel,
  author  = {Owhadi, Houman and Yoo, Gene Ryan},
  title   = {Kernel flows: from learning kernels from data into the abyss},
  journal = {Journal of Computational Physics},
  volume  = {389},
  pages   = {22--47},
  year    = {2019}
}

@article{yoo2021deep,
  author  = {Yoo, Gene Ryan and Owhadi, Houman},
  title   = {Deep regularization and direct training of the inner layers of neural networks with kernel flows},
  journal = {Physica D: Nonlinear Phenomena},
  volume  = {426},
  pages   = {132952},
  year    = {2021}
}

@book{owhadi2019operator,
  author    = {Owhadi, Houman and Scovel, Clint},
  title     = {Operator-Adapted Wavelets, Fast Solvers, and Numerical Homogenization},
  publisher = {Cambridge University Press},
  year      = {2019}
}

@article{mora2025operator,
  author  = {Mora, Carlos and Yousefpour, Amin and Hosseinmardi, Shirin and Owhadi, Houman and Bostanabad, Ramin},
  title   = {Operator learning with {Gaussian} processes},
  journal = {Computer Methods in Applied Mechanics and Engineering},
  volume  = {434},
  pages   = {117581},
  year    = {2025}
}

@book{wendland2004scattered,
  author    = {Wendland, Holger},
  title     = {Scattered Data Approximation},
  publisher = {Cambridge University Press},
  year      = {2004}
}

@article{marchenko1967distribution,
  author  = {Mar{\v{c}}enko, Vladimir A. and Pastur, Leonid A.},
  title   = {Distribution of eigenvalues for some sets of random matrices},
  journal = {Mathematics of the USSR-Sbornik},
  volume  = {1},
  number  = {4},
  pages   = {457--483},
  year    = {1967}
}

@article{baik2005phase,
  author  = {Baik, Jinho and Ben Arous, G{\'e}rard and P{\'e}ch{\'e}, Sandrine},
  title   = {Phase transition of the largest eigenvalue for nonnull complex sample covariance matrices},
  journal = {Annals of Probability},
  volume  = {33},
  number  = {5},
  pages   = {1643--1697},
  year    = {2005}
}

@inproceedings{vaswani2017attention,
  author    = {Vaswani, Ashish and Shazeer, Noam and Parmar, Niki and Uszkoreit, Jakob and Jones, Llion and Gomez, Aidan N. and Kaiser, {\L}ukasz and Polosukhin, Illia},
  title     = {Attention is all you need},
  booktitle = {Advances in Neural Information Processing Systems},
  volume    = {30},
  year      = {2017}
}

@inproceedings{dosovitskiy2021image,
  author    = {Dosovitskiy, Alexey and Beyer, Lucas and Kolesnikov, Alexander and Weissenborn, Dirk and Zhai, Xiaohua and Unterthiner, Thomas and Dehghani, Mostafa and Minderer, Matthias and Heigold, Georg and Gelly, Sylvain and Uszkoreit, Jakob and Houlsby, Neil},
  title     = {An image is worth 16x16 words: transformers for image recognition at scale},
  booktitle = {International Conference on Learning Representations},
  year      = {2021}
}

@article{caponnetto2007optimal,
  author  = {Caponnetto, Andrea and De Vito, Ernesto},
  title   = {Optimal rates for the regularized least-squares algorithm},
  journal = {Foundations of Computational Mathematics},
  volume  = {7},
  number  = {3},
  pages   = {331--368},
  year    = {2007}
}

@article{koltchinskii2000random,
  author  = {Koltchinskii, Vladimir and Gin{\'e}, Evarist},
  title   = {Random matrix approximation of spectra of integral operators},
  journal = {Bernoulli},
  volume  = {6},
  number  = {1},
  pages   = {113--167},
  year    = {2000}
}

@article{susiluoto2025radiative,
  author  = {Lamminp{\"a}{\"a}, Otto and Susiluoto, Jouni and Hobbs, Jonathan and McDuffie, James and Braverman, Amy and Owhadi, Houman},
  title   = {Forward model emulator for atmospheric radiative transfer using {Gaussian} processes and cross validation},
  journal = {Atmospheric Measurement Techniques},
  volume  = {18},
  pages   = {673--694},
  year    = {2025}
}

@misc{prasanth2021kernel,
  author       = {Prasanth, Sai and Haddad, Ziad S. and Susiluoto, Jouni and Braverman, Amy J. and Owhadi, Houman and Hamzi, Boumediene and Hristova-Veleva, Svetla M. and Turk, Joseph},
  title        = {Kernel flows to infer the structure of convective storms from satellite passive microwave observations},
  howpublished = {AGU Fall Meeting Abstracts, abstract A55F-1445},
  year         = {2021}
}

@article{wolpert1992stacked,
  author  = {Wolpert, David H.},
  title   = {Stacked generalization},
  journal = {Neural Networks},
  volume  = {5},
  number  = {2},
  pages   = {241--259},
  year    = {1992}
}

@inproceedings{yu2023climsim,
  author    = {Yu, Sungduk and Hannah, Walter M. and Peng, Liran and Lin, Jerry and Bhouri, Mohamed Aziz and Gupta, Ritwik and others},
  title     = {{ClimSim}: A large multi-scale dataset for hybrid physics-{ML} climate emulation},
  booktitle = {Advances in Neural Information Processing Systems (Datasets and Benchmarks Track)},
  volume    = {36},
  year      = {2023}
}

@article{darcy2023one,
  author  = {Darcy, Matthieu and Hamzi, Boumediene and Livieri, Giulia and Owhadi, Houman and Tavallali, Peyman},
  title   = {One-shot learning of stochastic differential equations with data adapted kernels},
  journal = {Physica D: Nonlinear Phenomena},
  volume  = {444},
  pages   = {133583},
  year    = {2023}
}

@book{vovk2005algorithmic,
  author    = {Vovk, Vladimir and Gammerman, Alexander and Shafer, Glenn},
  title     = {Algorithmic Learning in a Random World},
  publisher = {Springer},
  year      = {2005}
}

@article{lei2018distribution,
  author  = {Lei, Jing and G'Sell, Max and Rinaldo, Alessandro and Tibshirani, Ryan J. and Wasserman, Larry},
  title   = {Distribution-free predictive inference for regression},
  journal = {Journal of the American Statistical Association},
  volume  = {113},
  number  = {523},
  pages   = {1094--1111},
  year    = {2018}
}

\appendix
\section{Proofs}
\label{app:proofs}
\subsection{Proof of Proposition~\ref{prop:sym}}

By convexity of $\ell(\cdot,v)$,
\[
\ell\big(\widehat G_S(u),G(u)\big)\;\le\;\tfrac12\,\ell\big(\widehat
G(u),G(u)\big)+\tfrac12\,\ell\big(T\widehat G(Su),G(u)\big).
\]
Since $T$ is an involution, the equivariance $G(Su)=TG(u)$ applied at $u$
gives $TG(Su)=T^2G(u)=G(u)$, hence
\[
\ell\big(T\widehat G(Su),G(u)\big)=\ell\big(T\widehat G(Su),TG(Su)\big)
=\ell\big(\widehat G(Su),G(Su)\big),
\]
using the $T$-invariance of the loss. Taking expectations and using the
$S$-invariance of $\mu$ (so that $Su\sim\mu$ when $u\sim\mu$),
\[
\mathbb E\,\ell\big(\widehat G_S(u),G(u)\big)\le\tfrac12\,\mathbb
E\,\ell\big(\widehat G(u),G(u)\big)+\tfrac12\,\mathbb E\,\ell\big(\widehat
G(Su),G(Su)\big)=\mathbb E\,\ell\big(\widehat G(u),G(u)\big). \qed
\]

\subsection{Proof of Proposition~\ref{prop:kf}}

Strict positive definiteness of the restricted kernel matrix makes $I_C$ the
(unique) interpolant of the pairs indexed by $C$, so $I_C(z_j)=v_j$ for $j\in
C$ and the terms of \eqref{eq:kf} indexed by $C$ vanish, which is the first
claim. For the second, write
\[
e_2(\theta;C)\;=\;\sum_{i\in B\setminus C} g(C,i),\qquad
g(C,i):=\|v_i-I_C(z_i)\|_2^2 .
\]
Conditionally on $C$, the sum has exactly $b/2$ terms, so
$e_2(\theta;C)=\tfrac b2\,\mathbb E_{i\sim\mathrm{Unif}(B\setminus C)}[g(C,i)]$,
and taking the expectation over $C$ proves the identity. The gradient claim
is immediate: for fixed $(B,C)$ the map $\theta\mapsto e_2(\theta;C)$ is
differentiable wherever the Cholesky factorization in $I_C$ is (the kernel
matrix stays positive definite in a neighborhood), and under a local
integrable Lipschitz bound the derivative passes under the expectation, so
$\mathbb E_{B,C}[\nabla_\theta e_2]=\nabla_\theta\,\mathbb E_{B,C}[e_2]$. \qed

\subsection{Proof of Proposition~\ref{prop:stack}}

(i) Since $\sum_m w_m=1$,
$F_w(u)-G(u)=\sum_m w_m\,(f_m(u)-G(u))$, and the triangle inequality gives
$\|F_w(u)-G(u)\|_2\le\sum_m w_m\|f_m(u)-G(u)\|_2$. Dividing by $\|G(u)\|_2$
yields the pointwise claim; taking expectations gives the risk inequality, and
bounding each $\ell(f_m(u),G(u))$ by $B$ gives $\ell(F_w(u),G(u))\le B$.

(ii) Write $\ell_u(w)=\ell(F_w(u),G(u))$. First, $\ell_u$ is Lipschitz on
$\Delta$ for the $\ell^1$ norm: for $w,w'\in\Delta$, the reverse triangle
inequality and the argument of (i) give
\[
|\ell_u(w)-\ell_u(w')|\;\le\;\frac{\big\|\sum_m (w_m-w'_m)(f_m(u)-G(u))\big\|_2}
{\|G(u)\|_2}\;\le\;B\,\|w-w'\|_1 .
\]
Next, discretize the simplex. For $k\in\mathbb N$ let
$\mathcal G_k=\{w\in\Delta:\,kw\in\mathbb Z^M\}$; its cardinality is the number
of compositions of $k$ into $M$ nonnegative parts,
$\binom{k+M-1}{M-1}\le (k+1)^{M-1}$. Given $w\in\Delta$, set
$w'_i=\lfloor kw_i\rfloor/k$ for $i<M$ and $w'_M=1-\sum_{i<M}w'_i$; then
$w'\in\mathcal G_k$ (the last coordinate is a multiple of $1/k$ and is
$\ge w_M\ge0$ because the first $M-1$ coordinates only decreased), and
\[
\|w-w'\|_1=\sum_{i<M}(w_i-w'_i)+\big(w'_M-w_M\big)
=2\sum_{i<M}(w_i-w'_i)\;\le\;\frac{2(M-1)}{k}.
\]
By (i) the per-sample loss lies in $[0,B]$, so for each fixed $w'$ Hoeffding's
inequality gives
$\mathbb P\big(|\widehat R(w')-R(w')|>t\big)\le2e^{-2mt^2/B^2}$, and a union
bound over $\mathcal G_k$ shows that with probability at least $1-\delta$,
\[
\sup_{w'\in\mathcal G_k}\big|\widehat R(w')-R(w')\big|\;\le\;
B\sqrt{\frac{\log\!\big(2|\mathcal G_k|/\delta\big)}{2m}} .
\]
On this event, for every $w\in\Delta$, approximating by its grid point and
using the Lipschitz property for both $R$ and $\widehat R$,
\[
\big|\widehat R(w)-R(w)\big|\;\le\;
B\sqrt{\frac{\log(2|\mathcal G_k|/\delta)}{2m}}+\frac{4B(M-1)}{k}.
\]
If $w^\star$ minimizes $R$ over $\Delta$, the empirical minimizer satisfies
$R(F_{\widehat w})\le\widehat R(\widehat w)+\sup_\Delta|\widehat R-R|
\le\widehat R(w^\star)+\sup_\Delta|\widehat R-R|
\le R(F_{w^\star})+2\sup_\Delta|\widehat R-R|$. Choosing $k=m(M-1)$ and using
$|\mathcal G_k|\le(m(M-1)+1)^{M-1}$ gives
\[
R(F_{\widehat w})\;\le\;\min_{w\in\Delta}R(F_w)+\frac{8B}{m}
+B\sqrt{\frac{2\big((M-1)\log(m(M-1)+1)+\log(2/\delta)\big)}{m}} ,
\]
which is the claim. \qed

\subsection{Proof of Proposition~\ref{prop:secmom}}

Since $\sum_m w_m=1$, the stack's normalized residual is
$\rho_{f_w}(u)=\sum_m w_m\rho_{f_m}(u)$, hence
\[
\mathbb E\,\ell\big(f_w(u),G(u)\big)^2
=\mathbb E\Big\|\sum_m w_m\rho_{f_m}(u)\Big\|_2^2
=\sum_{m,k}w_mw_k\,\mathbb E\,\langle\rho_{f_m}(u),\rho_{f_k}(u)\rangle
=w^\top Sw .
\]
For (i), parameterize $w=(1-t,t)$; $q(t)=w^\top Sw$ is a convex quadratic in
$t$ with $q'(0)=2(\varrho e_1e_2-e_1^2)$, negative precisely when
$\varrho<e_1/e_2$. In that case the unconstrained minimizer lies in $(0,1)$ and
gives the stated interior value; otherwise $q'(0)\ge0$, the minimum over
$[0,1]$ is at $t=0$ with value $e_1^2$, and the second member is dropped. For
(ii), $S=e^2\big((1-\varrho)I+\varrho\,\mathbf1\mathbf1^\top\big)$, so
$w^\top Sw=e^2\big((1-\varrho)\|w\|_2^2+\varrho\big)$ on the simplex, minimized
by the uniform weights, where $\|w\|_2^2=1/M$. \qed

\subsection{Proof of Corollary~\ref{cor:floor}}

Convex weights are nonnegative, so each term of
$w^\top Sw=\sum_{m}w_m^2S_{mm}+\sum_{m\neq k}w_mw_kS_{mk}$ can be bounded
below entrywise:
\[
w^\top Sw\;\ge\;\bar e^{\,2}\sum_m w_m^2+\bar\varrho\,\bar
e^{\,2}\sum_{m\neq k}w_mw_k
=\bar e^{\,2}\Big(\|w\|_2^2+\bar\varrho\,(1-\|w\|_2^2)\Big),
\]
using $\sum_{m\neq k}w_mw_k=(\sum_m w_m)^2-\|w\|_2^2=1-\|w\|_2^2$. The bracket
is a convex combination of $1$ and $\bar\varrho$, hence at least
$\bar\varrho$. \qed

\subsection{Proof of Proposition~\ref{prop:percoord}}

The metric is diagonal, so the squared norm splits over coordinates and the
$j$-th summand $w_j^2\,\mathbb E\big(\sum_m v_{mj}f_m(u)_j-G(u)_j\big)^2$
depends on the weights only through the column $v_{\cdot j}$. Minimizing the
sum is therefore $q$ independent minimizations, and the positive factor
$w_j^2$ does not move any of the $q$ argmins, so the optimizer is the same
for every positive $w$. A combination with shared weights is the special
case $v_{mj}=v_m$ for all $j$, a subset of the feasible set of each
coordinate problem, which gives the domination. \qed

\subsection{Proof of Proposition~\ref{prop:pullback}}

Let $T:\mathcal H_k\to\mathbb R^{\mathcal X}$ be the composition map
$Tf=f\circ\varphi$. For $x\in\mathcal X$ the reproducing property gives
$(Tf)(x)=f(\varphi(x))=\langle f,k(\cdot,\varphi(x))\rangle_{\mathcal H_k}$,
so evaluation of $Tf$ at $x$ is a bounded functional, and the image
$\mathcal H_{k_\varphi}:=T(\mathcal H_k)$ carries the quotient norm
$\|g\|_{\mathcal H_{k_\varphi}}=\min\{\|f\|_{\mathcal H_k}:Tf=g\}$, the minimum
over the affine subspace $T^{-1}(g)$ (nonempty exactly when $g$ is a pullback).
This normed space is an RKHS: its reproducing kernel is
$k_\varphi(x,x')=\langle k(\cdot,\varphi(x)),k(\cdot,\varphi(x'))\rangle_{\mathcal H_k}
=k(\varphi(x),\varphi(x'))$, since the minimum-norm representer of evaluation
at $x$ is $k(\cdot,\varphi(x))$. Taking $f=h$ in the minimum gives
$\|h\circ\varphi\|_{\mathcal H_{k_\varphi}}\le\|h\|_{\mathcal H_k}$, and
substituting this bound into Theorem~\ref{thm:or} applied with the kernel
$k_\varphi$ replaces $\|G\|$ by $\|h\|_{\mathcal H_k}$. \qed

\subsection{Proof of Proposition~\ref{prop:aniso}}

Both displays are the native-space estimate for kernel ridge regression: for a
Mat\'ern kernel of smoothness $s$ on a bounded domain, with the nugget chosen
as in the pipeline, an estimator $\widehat G_n$ of a target $F$ in the native
space obeys $\mathbb E\|F-\widehat G_n\|\le C\,h_X^{\,s}\,\|F\|_{\mathcal N}$,
with $h_X$ the fill distance of the design in the kernel's metric
\citep{wendland2004scattered}, and for $n$ samples from a density bounded
above and below on a set of effective dimension $d'$, quasi-uniformity gives
$h_X\asymp(\log n/n)^{1/d'}$ with high probability, which we write as
$n^{-1/d'}$ up to the logarithmic factor absorbed into the statement.

For the isotropic kernel the metric is Euclidean on the full $d$-dimensional
domain, so $h_X\asymp n^{-1/d}$; the target is $F=g(A\,\cdot)$, whose Mat\'ern
native norm is equivalent to its $H^{s}$ norm. Changing variables $y=Au$ gives
$\|g(A\cdot)\|_{H^{s}}\asymp|\det A|^{-1/2}\,\sigma_1^{\,s}$, since each
derivative of order up to $s$ brings down at most a factor $\sigma_1$ and the
Jacobian contributes $|\det A|^{-1/2}$ in $L^2$; the $|\det A|^{-1/2}$ is one
of the singular-value constants absorbed into the statement, leaving the
displayed $\sigma_1^{\,s}$. For the adapted kernel $k_A(u,u')=k(Au,Au')$ the
estimator is the isotropic estimator of the unit-norm $g$ on the design
$\{Au_i\}$. Here the approximate-rank hypothesis enters: the design
$A\,\mathcal U$ has extent $\sigma_{r+1}\lesssim n^{-1/r}$ in each of the
trailing $d-r$ directions, below the fill distance those directions would
otherwise demand, so at resolution $n^{-1/r}$ the design is $r$-dimensional and
its fill distance is $h_X\asymp n^{-1/r}$ (a further $(\prod_{i\le r}\sigma_i)$
constant, again absorbed). Substituting the two fill distances into the
native-space estimate and dividing gives the ratio. If instead $A$ is close to
full rank, $\sigma_{r+1}$ is not below $n^{-1/r}$, the trailing directions are
resolved, and the adapted fill distance is $n^{-1/d}$ like the isotropic one:
the rates coincide and only $\sigma_1^{\,s}$ separates the bounds. \qed

\subsection{Proof of Theorem~\ref{thm:or}}

Fix $u$ and write $\kappa=k(X,u)\in\mathbb R^n$, $A=K+n\lambda I$,
$w=A^{-1}\kappa$. For each component $j$, membership $r_j\in\mathcal H_k$ and
the reproducing property give
\[
r_j(u)-\widehat r_{\lambda,j}(u)
= \Big\langle r_j,\; k(u,\cdot)-\sum_{i=1}^n w_i\,k(u_i,\cdot)\Big\rangle_{\mathcal H_k},
\]
because $\widehat r_{\lambda,j}(u)=\kappa^\top A^{-1}r_j(X)=\sum_i w_i\,r_j(u_i)$
and $r_j(u_i)=\langle r_j,k(u_i,\cdot)\rangle$. Cauchy--Schwarz yields
\[
\big|r_j(u)-\widehat r_{\lambda,j}(u)\big|\;\le\;\|r_j\|_{\mathcal H_k}\,
\Big\| k(u,\cdot)-\sum_i w_i k(u_i,\cdot)\Big\|_{\mathcal H_k},
\]
and the second factor is independent of $j$; call it $\rho(u)$. Expanding,
\[
\rho(u)^2 = k(u,u) - 2\,w^\top\kappa + w^\top K w
= k(u,u) - \kappa^\top A^{-1}\big(2A-K\big)A^{-1}\kappa .
\]
Since $2A-K=A+n\lambda I$,
\[
\rho(u)^2 = k(u,u)-\kappa^\top A^{-1}\kappa - n\lambda\,\|A^{-1}\kappa\|_2^2
=\widetilde P_\lambda(u)^2\;\le\;P_\lambda(u)^2 .
\]
Summing the squared componentwise bounds,
\[
\big\|r(u)-\widehat r_\lambda(u)\big\|_2^2=\sum_j\big(r_j(u)-\widehat
r_{\lambda,j}(u)\big)^2\le\rho(u)^2\sum_j\|r_j\|^2_{\mathcal H_k}
=\widetilde P_\lambda(u)^2\,\|r\|_K^2 . \qed
\]

Note that the argument does not use interpolation: it holds for every
$\lambda\ge0$, with the (slightly sharper) factor $\widetilde P_\lambda$
showing that smoothing can only tighten this particular bound relative to the
posterior standard deviation $P_\lambda$ that we report.

\subsection{Proof of Proposition~\ref{prop:conf}}

Write $s_i=\|e(\tilde u_i)\|_2/P_\lambda(\tilde u_i)$ for the validation scores
and $s^\star=\|e(u^\star)\|_2/P_\lambda(u^\star)$ for the test score. Because
$m$, $\widehat r_\lambda$, $X$ and $P_\lambda$ are fixed and the inputs
$\tilde u_1,\dots,\tilde u_m,u^\star$ are exchangeable and independent of them,
the scores $s_1,\dots,s_m,s^\star$ are exchangeable. The event
$G(u^\star)\in C(u^\star)$ is exactly $\{s^\star\le q\}$, where
$q=s_{(\lceil(1-\alpha)(m+1)\rceil)}$ is the
$k:=\lceil(1-\alpha)(m+1)\rceil$-th order statistic of the validation scores.

Consider the augmented sample $s_1,\dots,s_m,s^\star$ of size $m+1$ and let
$\mathrm{rank}(s^\star)$ be its rank (ties broken uniformly at random, which
only helps). By exchangeability the rank is uniform on $\{1,\dots,m+1\}$, so
$\mathbb P(\mathrm{rank}(s^\star)\le k)=k/(m+1)$. If $s^\star$ is among the $k$
smallest of the $m+1$ values then at most $k-1$ of the $s_i$ are below it, so
$s^\star\le s_{(k)}=q$; conversely $s^\star\le q$ forces
$\mathrm{rank}(s^\star)\le k$ except possibly through ties. Hence
\[
\mathbb P(s^\star\le q)\;\ge\;\mathbb P(\mathrm{rank}(s^\star)\le k)
=\frac{k}{m+1}=\frac{\lceil(1-\alpha)(m+1)\rceil}{m+1}\;\ge\;1-\alpha,
\]
which is the lower bound. When the scores are almost surely distinct there are
no ties, $\{s^\star\le q\}=\{\mathrm{rank}(s^\star)\le k\}$ exactly, and
$k/(m+1)<((1-\alpha)(m+1)+1)/(m+1)=1-\alpha+1/(m+1)$, giving the upper bound.
\qed

\subsection{Proof of Corollary~\ref{cor:mp}}

The nonzero eigenvalues of $K/n=XX^\top\!/n$ coincide with those of the
sample covariance matrix $S=X^\top X/n\in\mathbb R^{p\times p}$, and zero
eigenvalues contribute nothing to
$d_{\mathrm{eff}}(\lambda)=\sum_i\lambda_i(K)/(\lambda_i(K)+n\lambda)$, so
\[
\frac{d_{\mathrm{eff}}(\lambda)}{n}
=\frac{p}{n}\cdot\frac1p\sum_{j=1}^{p}\frac{\lambda_j(S)}{\lambda_j(S)+\lambda}
=\frac{p}{n}\Big(1-\lambda\,\widehat m_S(-\lambda)\Big),
\]
by the computation of Lemma~\ref{lem:deff} applied to $S$, with
$\widehat m_S$ the Stieltjes transform of the empirical spectral distribution
of $S$. By the Mar\v{c}enko--Pastur theorem
\citep{marchenko1967distribution}, this distribution converges weakly, almost
surely, to the law with ratio $\gamma$ and scale $\sigma^2$, and since
$x\mapsto(x+\lambda)^{-1}$ is bounded and continuous on $[0,\infty)$,
$\widehat m_S(-\lambda)\to m(-\lambda)$. It remains to evaluate
$m(-\lambda)$. The Stieltjes transform of the Mar\v{c}enko--Pastur law
satisfies the self-consistent equation
\[
m(z)=\frac{1}{\sigma^2\big(1-\gamma-\gamma z\,m(z)\big)-z},
\]
i.e.\ $\gamma\sigma^2 z\,m^2+\big(z-\sigma^2(1-\gamma)\big)m+1=0$. At
$z=-\lambda$ the discriminant is
$\big(\lambda+\sigma^2(1-\gamma)\big)^2+4\gamma\sigma^2\lambda>0$, and of the
two real roots
\[
m(-\lambda)=\frac{\sigma^2(1-\gamma)+\lambda\pm
\sqrt{\big(\lambda+\sigma^2(1-\gamma)\big)^2+4\gamma\sigma^2\lambda}}
{-2\gamma\sigma^2\lambda}
\]
only the one with the minus sign in the numerator is positive, as
$m(-\lambda)=\int(x+\lambda)^{-1}dF(x)>0$ requires; rearranging gives the
stated expression. \qed

\subsection{Proof of Lemma~\ref{lem:deff}}

Diagonalize $K$ with eigenvalues $\lambda_i$. Then
\[
\operatorname{tr}\big(K(K+n\lambda I)^{-1}\big)
=\sum_{i=1}^n\frac{\lambda_i}{\lambda_i+n\lambda}
=\sum_{i=1}^n\Big(1-\frac{n\lambda}{\lambda_i+n\lambda}\Big)
= n-\lambda\sum_{i=1}^n\frac{1}{\lambda_i/n+\lambda}
= n\big(1-\lambda\,\widehat m(-\lambda)\big),
\]
using $\widehat m(-\lambda)=\tfrac1n\sum_i(\lambda_i/n+\lambda)^{-1}$. The
limit statement follows from the weak convergence of the empirical spectral
distribution together with the boundedness and continuity of
$x\mapsto(x+\lambda)^{-1}$ on $[0,\infty)$ for fixed $\lambda>0$. \qed

\section{Implementation details}
\label{app:impl}
All experiments ran on a single laptop (one NVIDIA RTX~PRO~2000, 8~GB; 16 CPU
cores, 64~GB RAM), in single precision for network training and double
precision for all kernel solves and error computations. The dataset is the
distributed \texttt{StructuralMechanics} file pair (40000 samples); inputs are
reduced to $\mathbb R^{41}$ after verifying the broadcast structure exactly.
Splits: the training block is samples 1--20000, the test set is samples
20001--40000. High-data protocol: a fixed permutation of the training block
reserves 1000 samples for validation, 19000 for fitting. Low-data protocol:
the first 1250 samples of the training block, of which the last 250 form the
validation split. The test set is never touched during development; each
configuration is evaluated on it once, after selection on validation.

\paragraph{FNO.} Width 64, 14 modes per dimension, 4 spectral layers with
pointwise linear skips and GELU, lift from 3 channels (broadcast load, two
coordinates), projection $64\to128\to1$. AdamW, learning rate
$1.5\times10^{-3}$ ($2\times10^{-3}$ with batch 256), weight decay $10^{-6}$,
cosine schedule to $10^{-6}$, 300 epochs, batch 256. 6.45M parameters.

\paragraph{Transformer (implemented, not in the reported ensemble).} Encoder:
41 tokens (load value and 10 sine/cosine pairs of the coordinate), 5 pre-norm
blocks, dimension 192, 4 heads, MLP ratio 4. Decoder: grid queries from
Fourier features of both coordinates, two cross-attention blocks, pointwise
head $192\to192\to1$; 3.16M parameters. The cross-attention over 1681 grid
queries is the most compute-intensive of our means, and the hardware available
for this study (a single laptop GPU that proved unstable under sustained load)
did not let us train it to the level of the other members; we therefore
exclude it from the reported ensemble and note it as the natural fourth
architecture family for a follow-up on stable hardware.

\paragraph{UNet.} Three convolutional scales ($41\to21\to11$ by average
pooling, ceil mode) and a bottleneck at $6\times6$; two $3\times3$
convolutions with GroupNorm(8) and SiLU per block; widths $48/96/192/384$;
bilinear upsampling with skip concatenation; $1\times1$ output head. Input
channels as for the FNO. AdamW, learning rate $1.5\times10^{-3}$, weight
decay $10^{-5}$, cosine schedule, 200 epochs, batch 256. 4.38M parameters.

\paragraph{MSE variant.} The residual MLP trained with mean squared error on
standardized targets in place of the metric loss, otherwise identical recipe;
it enters the ensemble as a loss-diversity member and is also the strongest
plain MLP we obtain.

\paragraph{MLP.} Input layer $41\to1024$, three residual SiLU blocks of width
1024, output $1024\to1681$. AdamW, learning rate $10^{-3}$, weight decay
$10^{-5}$, cosine schedule, 400 epochs, batch 256. 4.91M parameters. A wider
variant (width 1536, five residual blocks, 14.5M parameters) is trained under
the same recipe.

\paragraph{Refiner.} The same residual trunk with input layer
$(41+1681)\to1024$: the load is concatenated with the kernel method's stress
prediction for that load. Training uses four-fold out-of-fold kernel
predictions for the input channel and full-data kernel predictions at
evaluation; otherwise identical to the MLP. 6.64M parameters. Reflection
augmentation flips the load and permutes both the kernel channel and the target
by the row-reversal index.

\paragraph{Common training details.} Loss: mean over the batch of
$\|\hat v-v\|_2/\|v\|_2$ in original units (predictions denormalized inside
the loss; targets standardized by the per-pixel training mean and global
standard deviation), except for the MSE variant above. Reflection
augmentation with probability $1/2$; reflection averaging at evaluation.
Model selection: validation error computed every 10 epochs, best checkpoint
kept. A single seed is trained per architecture; the ensemble's diversity
comes from the architectures and losses rather than from reseeding, and the
correlation analysis of Section~\ref{sec:diversity} indicates same-architecture
reseeds would be more correlated still.

\paragraph{Stacking.} Global weights: convex, initialized at the simplex
minimizer of the measured second-moment matrix $S$
(Proposition~\ref{prop:secmom}) and polished by a short random search on the
validation metric. Per-pixel weights: affine in the members at each grid
point, ridge parameter $10^{-3}$, fitted on half the validation split and
accepted only if they beat the global weights on the held-out half, then
refitted on the full split.

\paragraph{Kernel stages.} Mat\'ern-$5/2$ on standardized inputs. Scale grid
$\{0.5,1,2,4\}\times$ the median pairwise distance (estimated on 2000
points), nugget grid $\{10^{-7},10^{-5},10^{-3}\}$ (scaled by $n$), tuned on
validation using an 8000-sample subsample of the training set, refit at the
chosen pair on the full training set in double precision. The pure kernel
baseline (Table~\ref{tab:main}) uses the grid
$\{0.5,0.75,1,1.5,2\}\times$median and nuggets $\{10^{-8},10^{-6},10^{-4}\}$
directly at $n=19000$. Feature stage: penultimate activations of the ensemble
members (FNO: spatially averaged pre-projection channels, 128; transformer:
mean-pooled encoder tokens, 192; MLP: trunk output, 1024), concatenated,
standardized, same kernel family and tuning.

\paragraph{Uncertainty.} Posterior standard deviation \eqref{eq:power}
computed from the Cholesky factor of $K+n\lambda I$ by triangular solves.
Conformal scaling: the 0.9-quantile of $\|e\|_2/P_\lambda$ on the validation
split multiplies $P_\lambda$ on test; coverage is the fraction of test
samples whose absolute error norm falls below the scaled band.

\end{document}